\documentclass[lettersize,journal]{IEEEtran}
\usepackage{amsmath,amsfonts}
\usepackage{algorithm}
\usepackage{algpseudocode}
\usepackage{array}
\usepackage[caption=false,font=normalsize,labelfont=sf,textfont=sf]{subfig}
\usepackage{textcomp}
\usepackage{stfloats}
\usepackage{url}
\usepackage{verbatim}
\usepackage{graphicx}
\usepackage{cite}
\usepackage{graphicx}  % To include graphics
\usepackage{verbatim}  % To display text as-is
\usepackage{booktabs}  % To create better-looking table rules
\usepackage{multirow}  % To merge rows in a table
\usepackage{makecell}  % For flexible formatting within table cells
\usepackage{amssymb}  % To access additional math symbols
\usepackage{lipsum}  % To generate dummy text for testing or demonstration
\usepackage{colortbl}  % To color table cells
\usepackage{soul}  % For highlighting and strikethrough effects on text
\usepackage{microtype}  % For fine-tuning text layout
\usepackage{bm}  % For bold math symbols and vectors
\usepackage{makecell}  % Again for formatting within table cells
\usepackage{multirow}  % Again for merging rows in a table
\usepackage{soul}  % Again for text effects
\usepackage{caption}  % To customize captions of figures and tables
\usepackage{enumitem}  % For customizing list environments
\usepackage{amsmath}  % For advanced math typesetting
\usepackage{float}  % For better control of floating objects like figures and tables
\usepackage[colorlinks]{hyperref}  % To add hyperlinks and color them
\usepackage{lineno}  % To add line numbers
\usepackage{array}  % For enhanced array and table formatting
\usepackage{xcolor}
\usepackage{caption}

\newcommand{\major}[1]{\textcolor{black}{#1}}

\begin{document}

%\title{Unsupervised Online 3D Instance Segmentation with Data Synthesis and Dynamic Loss}
\title{Unsupervised Online 3D Instance Segmentation with Synthetic Sequences and Dynamic Loss}

\author{
        Yifan Zhang, Wei Zhang, Chuangxin He, Zhonghua Miao,
        Junhui Hou,~\IEEEmembership{Senior Member,~IEEE},
        % IEEE Publication Technology,~\IEEEmembership{Staff,~IEEE,}
        % <-this % stops a space
\thanks{
% This work was supported in part by the NSFC Excellent Young Scientists Fund 62422118, and in part by the Hong Kong Research Grants Council under Grants 11219324 and 11219422. (Corresponding author: Junhui Hou)
This work was supported in part by the National Natural Science Foundation of China under Grants 62422118, 52375107 and 32401712, in part by the Hong Kong Research Grants Council under Grants  11219324, 11219422, and N\_CityU1114/25, in part by the Natural Science Foundation of Shanghai under Grant 24ZR1424800, and in part by the Shanghai Pujiang Program 25PJA042. (\textit{Corresponding author: Junhui Hou})
}
%\thanks{Manuscript received April 19, 2021; revised August 16, 2021.}
\thanks{Y. Zhang is with the School of Mechatronic Engineering and Automation, Shanghai University, Shanghai, China, and also with the Department of Computer Science, City University of Hong Kong, Hong Kong SAR, China. E-mail: yfzhang@shu.edu.cn;}
\thanks{W. Zhang, C. He, and Z. Miao are with the School of Mechatronic Engineering and Automation, Shanghai University, Shanghai, China. E-mail: {chuangxinhe@shu.edu.cn;wzhang23@shu.edu.cn;zhhmiao@shu.edu.cn;}}
\thanks{J. Hou is with the Department of Computer Science, City University of Hong Kong, Hong Kong SAR, China. E-mail: jh.hou@cityu.edu.hk.}
}

% The paper headers
\markboth{}%Journal of \LaTeX\ Class Files,~Vol.~14, No.~8, August~2021
{Shell \MakeLowercase{\textit{et al.}}: A Sample Article Using IEEEtran.cls for IEEE Journals}

% \IEEEpubid{0000--0000/00\$00.00~\copyright~2021 IEEE}
% Remember, if you use this you must call \IEEEpubidadjcol in the second
% column for its text to clear the IEEEpubid mark.

\maketitle

\begin{abstract}
Unsupervised online 3D instance segmentation is a fundamental yet challenging task, as it requires maintaining consistent object identities across LiDAR scans without relying on annotated training data. Existing methods, such as UNIT, have made progress in this direction but remain constrained by limited training diversity, rigid temporal sampling, and heavy dependence on noisy pseudo-labels. 
%	To address these issues, we introduce a novel framework that integrates three key innovations: (1) point cloud sequence synthesis, which generates diverse synthetic training sequences directly from unsupervised clustering results, enriching the training distribution without manual labels or simulation engines; (2) flexible temporal sampling, which captures both short- and long-range dependencies by leveraging adjacent and non-adjacent frames; and (3) a dynamic-weighting loss, which emphasizes confident and dynamic samples, guiding the network toward more informative instances. 
We propose a new framework that enriches the training distribution through synthetic point cloud sequence generation, enabling greater diversity without relying on manual labels or simulation engines. To better capture temporal dynamics, our method incorporates a flexible sampling strategy that leverages both adjacent and non-adjacent frames, allowing the model to learn from long-range dependencies as well as short-term variations. In addition, a dynamic-weighting loss emphasizes confident and informative samples, guiding the network toward more robust representations.
Through extensive experiments on SemanticKITTI, nuScenes, and PandaSet, our method consistently outperforms UNIT and other unsupervised baselines, achieving higher segmentation accuracy and more robust temporal associations. The code will be publicly available at \textbf{github.com/Eaphan/SFT3D}.
%	These results demonstrate the effectiveness and generalization ability of our approach, advancing the state of the art in unsupervised online 3D instance segmentation.
\end{abstract}


\section{Introduction}
\IEEEPARstart{T}{he} rapid advancement of 3D sensing technologies, particularly LiDAR, has opened up new possibilities for applications such as autonomous driving and robotics~\cite{wu2023leveraging,zhang2023pointmcd}. A key task in these applications is 3D instance segmentation, which involves identifying and delineating individual objects within a 3D point cloud~\cite{hou20193d,camuffo2023learning}. This task is particularly challenging due to the complexity of 3D data, the variability of object shapes and sizes, and the need for accurate real-time processing. Traditionally, instance segmentation relies on large, manually annotated datasets, which are both time-consuming and costly to produce~\cite{schult2022mask3d}. Moreover, many real-world scenarios require the ability to process data sequentially in an online manner, where data arrives continuously, and future information is unavailable. This presents a significant challenge for maintaining temporal consistency and robust instance tracking across consecutive 3D scans. Therefore, developing methods for unsupervised online 3D instance segmentation is essential~\cite{sautier2024unit,yi2025ov}.
%, as it can enable real-time, scalable, and label-efficient solutions for dynamic environments.

Several prior works have focused on unsupervised 3D instance segmentation. TARL-Seg~\cite{tarl} and 4D-Seg~\cite{sautier2024unit} are offline methods that rely on aggregating multiple consecutive LiDAR scans to generate pseudo-labels for object instances. TARL-Seg applies the HDBSCAN clustering algorithm to temporally accumulated scans, but is limited by its reliance on fixed time windows and the need for scan registration~\cite{tarl}. 4D-Seg, an extension of TARL, improves upon this by processing a longer temporal window and leveraging voxel grid sampling to reduce computational complexity, though it still suffers from similar limitations regarding temporal consistency and real-time processing. In contrast, 3DUIS is a scan-wise unsupervised method that performs instance segmentation by solving a graph-cut problem, but lacks the temporal consistency required for tracking objects over time~\cite{3duis}. UNIT, a more recent advancement, addresses the challenges of real-time instance segmentation by introducing an online framework~\cite{sautier2024unit}. Unlike offline methods, UNIT performs segmentation and tracking on individual scans in an auto-regressive manner, maintaining temporal consistency across scans without needing access to future data. This online capability makes UNIT more suitable for real-time applications, such as autonomous driving, where timely and continuous object tracking is essential.
% Suggestion: Do not mention the limitation of existing works, only describe the methds?

%Despite the progress made in unsupervised 3D instance segmentation, existing methods like UNIT still face several significant limitations. 
While existing methods, including UNIT, have made significant strides in unsupervised 3D instance segmentation, they still exhibit notable limitations.
Offline methods like TARL-Seg and 4D-Seg rely on the aggregation of multiple consecutive scans, which limits their applicability in real-time scenarios. These approaches require temporal accumulation and scan registration, both of which introduce additional computational complexity and reduce their efficiency in dynamic environments~\cite{nunes2022segcontrast,tarl}. Furthermore, these methods struggle with tracking objects over extended periods, as their segmentation is confined to predefined temporal windows~\cite{tarl}. On the other hand, scan-wise methods like 3DUIS lack temporal consistency and are unable to track objects across scans, making them unsuitable for applications where continuous tracking is critical~\cite{3duis}.

Although UNIT represents a significant advancement in unsupervised online 3D instance segmentation, it still faces several limitations that hinder its full potential. \textit{First}, UNIT lacks sufficient data augmentation, as it is heavily reliant on the diversity of the available training data. The model's performance is constrained by the limited variety present in the training set, and thus it struggles to generalize to new or unseen scenarios. To improve robustness, much greater diversity than what is present in the current data is essential. \textit{Second}, the training sample selection in UNIT is rather rigid, as it primarily uses pairs of adjacent and temporally ordered frames. This limited temporal context restricts the model's ability to capture long-range dependencies and dynamic changes in the scene, making it less effective in handling more complex temporal scenarios. \textit{Third}, UNIT heavily depends on the quality of the pseudo-labels generated through spatio-temporal clustering, which can introduce noise, especially in dynamic or cluttered environments. This reliance on noisy pseudo-labels often leads to segmentation errors, affecting the accuracy of instance tracking over time. \textit{Finally}, UNIT treats all samples equally, without considering the varying importance of different instances. In particular, dynamic and informative samples, which are crucial for accurate tracking and segmentation, are not given special emphasis during training. This uniform treatment of all samples may reduce the model's focus on more critical and motion-driven objects, thereby limiting its overall performance.

%To address these challenges, we propose several key innovations. First, we introduce point cloud sequence synthesis for training, which generates diverse synthetic sequences to augment training data, helping the model learn more robust features. Second, we propose flexible temporal sampling, allowing the model to capture long-range temporal dependencies and dynamic changes by sampling both adjacent and non-adjacent frames. Finally, we introduce a dynamic-weighting loss that prioritizes confident and dynamic instances, improving the model's focus on more informative training samples. These innovations aim to enhance the performance and generalization of the model, enabling more accurate and robust online instance segmentation in real-world, dynamic environments.

To overcome these challenges, we introduce several key innovations. First, we propose point cloud sequence synthesis for training, which generates diverse synthetic sequences to enrich the training data, thereby helping the model learn more robust and generalized features. Second, we introduce flexible temporal sampling, a strategy that allows the model to capture long-range temporal dependencies and adapt to dynamic changes by sampling both adjacent and non-adjacent frames. Finally, we incorporate a dynamic-weighting loss, which prioritizes confident and dynamic instances, ensuring that the model focuses more on informative training samples. These innovations collectively enhance the model's performance and generalization, enabling more accurate, robust, and scalable online instance segmentation in real-world, dynamic environments.

To evaluate our method, we conduct comprehensive experiments on three benchmark datasets: SemanticKITTI~\cite{behley2019semantickitti}, nuScenes~\cite{caesar2020nuscenes}, and PandaSet~\cite{xiao2021pandaset}. On the SemanticKITTI dataset, our approach achieves a temporal association score of 0.523, an $\mathrm{IoU^*}$ of 0.602, and an association score of 0.725, corresponding to consistent improvements over UNIT by +0.041, +0.034, and +0.029, respectively. Similar performance gains are also observed on the nuScenes and PandaSet datasets, where our method consistently outperforms the baseline across multiple metrics. These results clearly demonstrate the effectiveness of our approach.
To summarize, the main contributions of this work are as follows.
%\todo{To evaluate our method, we carry out comprehensive experiments on three datasets, SemanticKITTI, nuScenes, and PandaSet. 
%	On the SemanticKITTI dataset, our method achieves a temporal association score of 0.523, an $\mathrm{IoU^*}$ of 0.602, and an association score of 0.725, which correspond to consistent gains over UNIT of +0.041, +0.034, and +0.029, respectively. Our method also outperforms consistently the baseline on nuScenes and pandaset.
%	The results demonstrate that our method outperforms existing xxx approaches (证明我们方法的有效性?). } 
\begin{itemize}
	\item We introduce a novel method for generating diverse synthetic point cloud sequences, augmenting the training data to improve model robustness and generalization.
	\item We propose a flexible temporal sampling strategy that captures both short-term and long-term dependencies by selecting both adjacent and non-adjacent frames, enhancing the model's ability to track dynamic objects over time.
	\item We introduce a dynamic-weighting loss function that emphasizes dynamic and confident instances, ensuring that the model prioritizes the most informative and challenging samples during training.
\end{itemize}

%\todo{The rest of this paper is organized as follows. Section~\ref{sec:related_work} provides an overview of related literature pertinent to our research. In Section~\ref{sec:method}, we present the overall architecture of NCLR and elaborate its principal elements. Section~\ref{sec:experiments} presents an empirical evaluation of our proposed approach across three distinct downstream tasks, along with ablation studies to assess the impact of key components. The paper concludes with Section~\ref{sec:conclusion}, summarizing our findings.}

The remainder of this paper is organized as follows. Section~\ref{sec:related_work} reviews prior work pertinent to our research. Section~\ref{sec:method} introduces our proposed framework in detail. Section~\ref{sec:experiments} reports comprehensive evaluations on multiple datasets, along with ablation studies to assess the contributions of individual components. 
%\todo{Finally, Section~\ref{sec:conclusion} concludes the paper by summarizing our findings and discussing potential directions for future research.}
Finally, Section~\ref{sec:conclusion} concludes the paper by summarizing our findings.

\vspace{-0.2cm}
\section{Related Work}\label{sec:related_work}
\subsection{Unsupervised 3D Instance Segmentation} 
Unsupervised 3D instance segmentation has been studied in both indoor and outdoor settings. UnScene3D~\cite{rozenberszki2024unscene3d} addresses dense, static indoor point clouds by training Mask3D~\cite{schult2022mask3d} on pseudo-masks from self-supervised features, without requiring temporal consistency. For outdoor LiDAR, SegContrast~\cite{nunes2022segcontrast} generates pseudo-masks via ground removal and clustering, while TARL-Seg~\cite{tarl} extends this idea with spatio-temporal clustering over accumulated scans using Patchwork~\cite{patchwork} and HDBSCAN~\cite{hdbscan}. 3DUIS~\cite{3duis} applies graph-cut segmentation on SegContrast features in single scans, introducing the $\mathrm{S_{assoc}}$ metric, whereas AutoInst~\cite{perauer2024autoinst} leverages TARL~\cite{tarl} features (and optionally DINOv2~\cite{oquab2023dinov2}) on accumulated scans, further refining pseudo-labels with MaskPLS~\cite{marcuzzi2023mask}.

The first online approach, UNIT, employs an auto-regressive, query-based network to propagate instance identities across scans, trained with both instance mask and temporal consistency losses~\cite{sautier2024unit}. While effective, UNIT remains limited by the diversity of training data, rigid adjacent-frame sampling, and sensitivity to noisy pseudo-labels. Other works, such as OYSTER~\cite{zhang2023towards}, SeMoLi~\cite{seidenschwarz2024semoli}, and LISO~\cite{baur2024liso}, focus on discovering moving objects for unsupervised detection, but they target bounding boxes rather than full instance segmentation.

%\subsection{Point Cloud Sequence Synthesis}
\subsection{Synthetic Data Generation.}
Annotating large-scale datasets with precise labels is often labor-intensive and sometimes impractical. This challenge has driven a growing interest in generating synthetic training data to support diverse computer vision applications~\cite{wang2024videocutler,xiao2022polarmix,kong2023lasermix}. In the 3D object detection domain, a widely used augmentation strategy, often referred to as “GT-Aug”, involves sampling labeled objects from a database and inserting them into training scenes~\cite{zhang2023glenet,zhang2024stemd}. However, this straightforward copy-paste approach overlooks contextual placement and produces only limited variation in the generated samples~\cite{fang2021lidar}.
%To expand training data beyond static frame-wise augmentations, recent work generates point-cloud sequences that preserve spatiotemporal realism.
% Cross-scan composition methods such as PolarMix and LaserMix mix or paste real geometry across scans, while 
D-Aug advances the composition method to the sequence level by inserting objects consistently across consecutive frames with collision-aware, road-constrained placement~\cite{zhao2024d}. Complementarily, rendering/simulation frameworks (LiDAR-Aug~\cite{fang2021lidar}; LiDARsim~\cite{manivasagam2020lidarsim}/CARLA~\cite{dosovitskiy2017carla} pipelines) synthesize fully labeled LiDAR sequences with controllable dynamics and conditions, enabling rare-case and long-tail coverage. Emerging generative models further sample realistic scans—and increasingly, temporally coherent sequences—to enrich training beyond what composition or simulation alone can offer~\cite{zyrianov2022learning,xiang2024synthetic}. 
% Collectively, these approaches improve downstream detection, segmentation, and tracking by providing temporally consistent, diverse training trajectories.
%Unlike prior approaches that rely on manual labels, simulation environments, or simple copy-paste strategies, our method generates synthetic LiDAR point cloud sequences directly from unsupervised clustering results.
%Instead of inserting ground-truth objects or rendering scenes from scratch, we construct sequences by reusing pseudo-labeled object instances and placing them into valid ground regions with collision-aware constraints. This design ensures both spatial realism and temporal consistency, while requiring no human annotations or external simulators. In this way, our approach enriches the training distribution with diverse, physically plausible sequences tailored for online instance segmentation, complementing and extending the capabilities of existing data synthesis methods.
In contrast to prior approaches that depend on manual labels or simulation pipelines, our method constructs synthetic LiDAR sequences directly from unsupervised clustering results. By reusing pseudo-labeled object instances and placing them into valid ground regions with collision-aware constraints, we generate diverse and temporally consistent training data—without requiring annotations or external simulators.

\vspace{-0.1cm}
\subsection{3D Instance Segmentation} 
3D instance segmentation has evolved rapidly over the past few years, with methods broadly falling into proposal-based~\cite{hou20193d}, proposal-free (grouping/clustering), kernel-based~\cite{he2021dyco3d}, and transformer-based paradigms~\cite{schult2022mask3d,wang20233d}.
Proposal-based methods like 3D-SIS first detect object bounding boxes in 3D or via fused RGB-D input, then refine to instance masks, but these often depend strongly on the quality of the 3D detection step~\cite{hou20193d}. 
Grouping or proposal-free methods, such as SGPN~\cite{wang2018sgpn} / PointGroup~\cite{jiang2020pointgroup}, predict per-point features (embeddings, center offsets, or affinities) and then cluster or group them to form instances. For example, RPGN~\cite{dong2022learning} introduces a notion of regional purity to help mitigate over- and under‐segmentation in clustering. 
%More recently, transformer- and hybrid approaches seek to capture global context, with works like Insightful Instance Features for 3D Instance Segmentation categorizing different method families into proposal-based, grouping-based, kernel-based, and transformer-based, and showing improved boundary precision and semantic consistency. 
% Beyond closed-set instance segmentation, there is growing interest in open-vocabulary or open-world 3D instance segmentation, where methods such as OpenMask3D, Open3DIS, and Open-YOLO 3D (the latter recently proposed) attempt to recognize and segment novel object categories beyond those in the training set. Each of these threads addresses trade-offs among precision, computational cost, annotation cost, and generalization.
Mask3D~\cite{schult2022mask3d} is a Transformer-based model for 3D semantic instance segmentation that uses instance queries to directly predict instance masks from point clouds, eliminating the need for hand-crafted voting or grouping mechanisms.
%\major{Recent advancements in few-shot 3D point cloud segmentation (FS-PCS) have focused on improving generalization to novel classes. introduces COSeg, which addresses foreground leakage and sparse point distribution by refining correlations between query points and prototypes, enhancing generalization}
\major{COSeg~\cite{an2024rethinking} improves generalization by refining query-prototype correlations, while GFS-VL~\cite{an2025generalized} combines pseudo-labels from vision-language models with few-shot samples for better novel class segmentation. Seal~\cite{liu2023segment} distills vision foundation models (VFMs) for segmenting diverse automotive point cloud sequences.}

\begin{figure*}[t]
    \centering
    \includegraphics[width=0.9\textwidth]{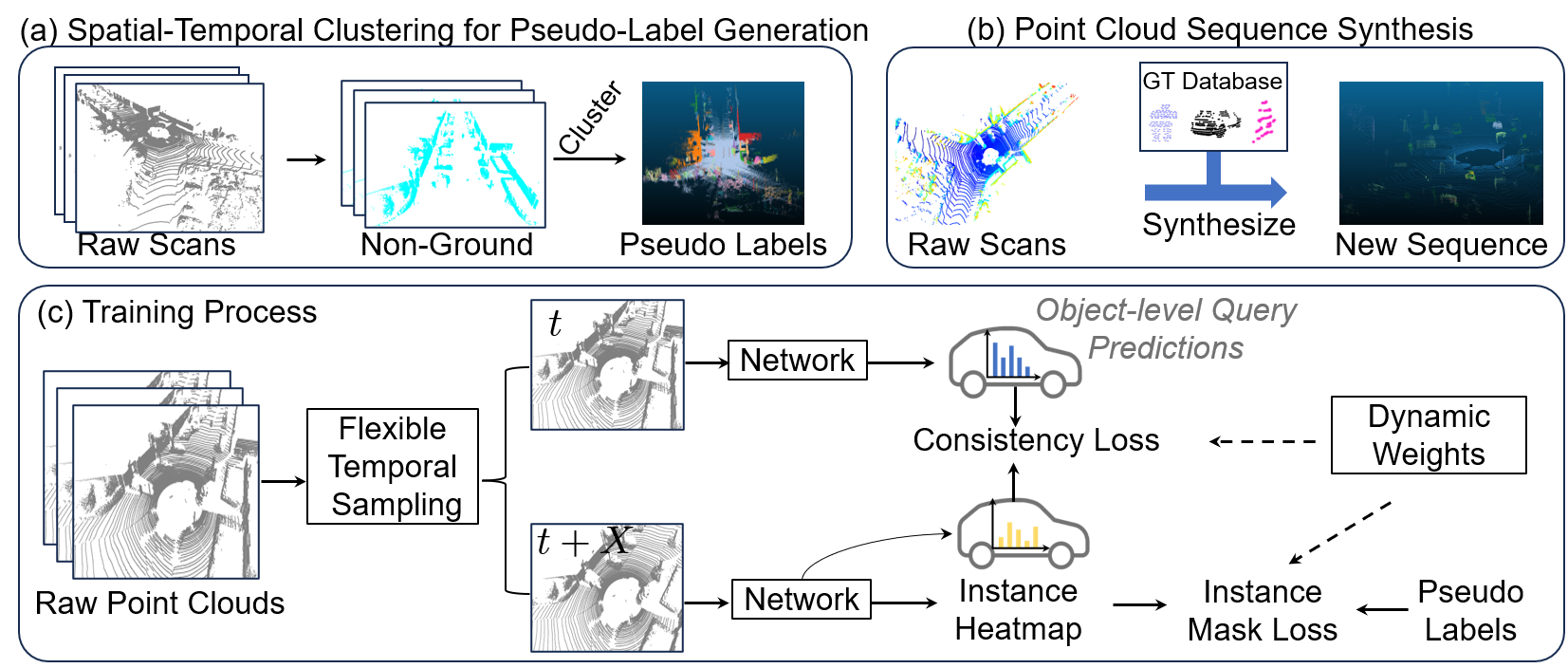} 
    \caption{
        The overall framework. (a) We obtain initial pseudo labels with spatial-temporal clustering. (b) We synthesize the new point cloud sequence for training. (c) We train the network with the pseudo-labels. 
        The samples are flexibly taken from a point cloud sequence, and the losses are adjusted with dynamic weights.
        %		We then use these pseudo-labels to train an auto-regressive network.
    }
    \label{fig:overview}
    \vspace{-0.18cm}
\end{figure*}

\vspace{-0.15cm}
\section{Proposed Method}\label{sec:method}
% 先用一副图来表示 整体的结构，然后介绍下各个模块
\subsection{Problem Formulation}
% What is the unsupervised online 3D instance segmentation?
%In recent years, autonomous systems have relied heavily on LiDAR sensors for accurate perception of the surrounding environment. One of the critical tasks in these systems is the segmentation and tracking of 3D objects from LiDAR point clouds, which forms the foundation for understanding dynamic scenes, such as vehicles, pedestrians, and static obstacles. This task, often referred to as **3D instance segmentation and tracking**, is essential for autonomous driving, robotics, and other applications requiring real-time spatial understanding.

The goal of this work is to tackle the unsupervised online 3D instance segmentation problem. Specifically, we aim to develop a method that performs instance segmentation without relying on manually annotated data, while also ensuring that object instances are tracked across time in an online fashion, i.e., processing data sequentially as it becomes available. This problem is challenging due to the lack of ground truth labels for instance segmentation, and the need to maintain temporal consistency over consecutive scans of the LiDAR point cloud.

Given a sequence of raw LiDAR scans (denoted as $\{P_t\}_{t=1}^T$), where each scan $P_t$ consists of a set of 3D points $\{(x_i, y_i, z_i)\}_{i=1}^{N_p^t}$ with associated features (such as intensity), the objective is to segment objects and track objects over time. Specifically, for each scan $P_t$, we aim to assign each point $p_i \in P_t$ to a unique instance, such that points belonging to the same object share the same instance ID, and different objects are assigned distinct instance IDs. Then we ensure temporal consistency by tracking the same object instance across consecutive scans. That is, the same object in $P_t$ and $P_{t+1}$ should be assigned the same instance ID, even if the object moves or changes appearance. This must be achieved in an online manner, without access to future scans.

\vspace{-0.15cm}
\subsection{Overview}
Figure~\ref{fig:overview} summarizes the full pipeline. Given unlabeled LiDAR sequences, we first derive class-agnostic pseudo-labels by spatio-temporal clustering, producing 4D instance segments that are consistent across scans (Fig. \ref{fig:overview}\textcolor{red}{(a)}). To enrich the training distribution, we then synthesize additional point-cloud sequences (Fig.~\ref{fig:overview}\textcolor{red}{(b)}).
%: ground points are extracted to form a BEV valid map, object point clouds are sampled from a database, and they are pasted into valid regions with collision checks to yield physically plausible trajectories.

The online segmenter is an auto-regressive, query-based network that consumes one scan at a time and propagates its query embeddings across timesteps to maintain identity over time (Fig. \ref{fig:overview}\textcolor{red}{(c)}). Training is fully unsupervised: matched query–instance masks are optimized with the instance mask loss, and temporal coherence is enforced with a consistency loss between object-level query predictions. To better capture dynamics, we adopt flexible temporal sampling that mixes adjacent and non-adjacent pairs and their reversed orders, and we stabilize learning with dynamic weighting—down-weighting low-confidence points and up-weighting instances that exhibit motion. In inference, the model operates purely online, without access to future scans or explicit multi-scan registration.

%\subsection{Pseudo Label Generation}
\subsection{Spatio-Temporal Clustering for Pseudo-Label Generation}\label{sec:clustering}

\textbf{Ground Segmentation.} Following prior work~\cite{nunes2022segcontrast,sautier2024unit}, we employ a ground segmentation strategy to isolate objects in LiDAR scans, which helps in differentiating objects from the background. As ground points typically form a large, mostly flat region, we can effectively separate them from the rest of the scene using methods such as RANSAC. However, more recent approaches~\cite{patchwork,patchworkpp} have been shown to outperform RANSAC by addressing the specific characteristics of LiDAR point clouds in an unsupervised manner. We utilize these advanced methods for ground segmentation. Specifically, we apply Patchwork~\cite{patchwork} to segment ground points in the SemanticKITTI dataset (as in TARL~\cite{tarl}), and Patchwork++~\cite{patchworkpp} for the nuScenes dataset. Both of these techniques operate scan-wise, processing each individual LiDAR scan in the training set.

%\noindent\textbf{Spatio-Temporal Segmentation.} Previous work by the authors of TARL~\cite{tarl} showed that applying HDBSCAN~\cite{hdbscan} to temporally accumulated scans enables the identification of object instances that are consistent both spatially and temporally, even for dynamic objects. However, the computational complexity of HDBSCAN limits its application to shorter time windows. In practice, TARL~\cite{tarl} aggregates 12 consecutive scans from the SemanticKITTI dataset, resulting in pseudo-labels over a temporal window of 1.2\,s, given the LiDAR’s operating frequency of 10\,Hz.\
%To address this limitation and extend the temporal context, we introduce voxel grid sampling to reduce the number of points prior to clustering after temporal aggregation. By employing this approach, we are able to generate temporally consistent clusters over a larger number of scans, up to 40 aggregated scans.
% As shown in \cref{sec:experiments} (\cref{table:results_sk}), the resulting clusters exhibit similar spatial accuracy to those generated in~\cite{tarl} when evaluated on a per-scan basis, but with significantly better temporal consistency.

\textbf{Spatio-Temporal Segmentation.}
Previous work by the authors of TARL~\cite{tarl} showed that applying HDBSCAN~\cite{hdbscan} to temporally accumulated scans enables the identification of object instances that are consistent both spatially and temporally, even for dynamic objects. However, the computational complexity of HDBSCAN limits its application to shorter time windows. 
%In practice, TARL~\cite{tarl} aggregates 12 consecutive scans from the SemanticKITTI dataset, resulting in pseudo-labels over a temporal window of 1.2\,s, given the LiDAR’s operating frequency of 10\,Hz.
Following the approach of UNIT~\cite{sautier2024unit}, we use voxel grid sampling to reduce the number of points to cluster after temporal aggregation. This technique helps extend the temporal context over a larger number of scans while keeping the computational burden manageable. Unlike TARL~\cite{tarl}, which aggregates 12 consecutive scans to form pseudo-labels over a 1.2\,s temporal window, this method allows for more flexibility and scalability by adapting to varying scan frequencies. 
%This results in temporally consistent object instance clustering, with improved efficiency and high spatial accuracy, as demonstrated in \cref{sec:experiments}.

\subsection{Point Cloud Sequence Synthesis for Training}
%Firstly, we generate pseudo-masks for multiple objects in an image using MaskCut [35]. Then, we convert a random pair of images in the minibatch into a video with corresponding pseudo mask trajectories using ImageCut2Video. Finally, we train an unsupervised video instance segmentation model using these mask trajectories.

\begin{figure}[t]
    \centering
    \includegraphics[width=0.47\textwidth]{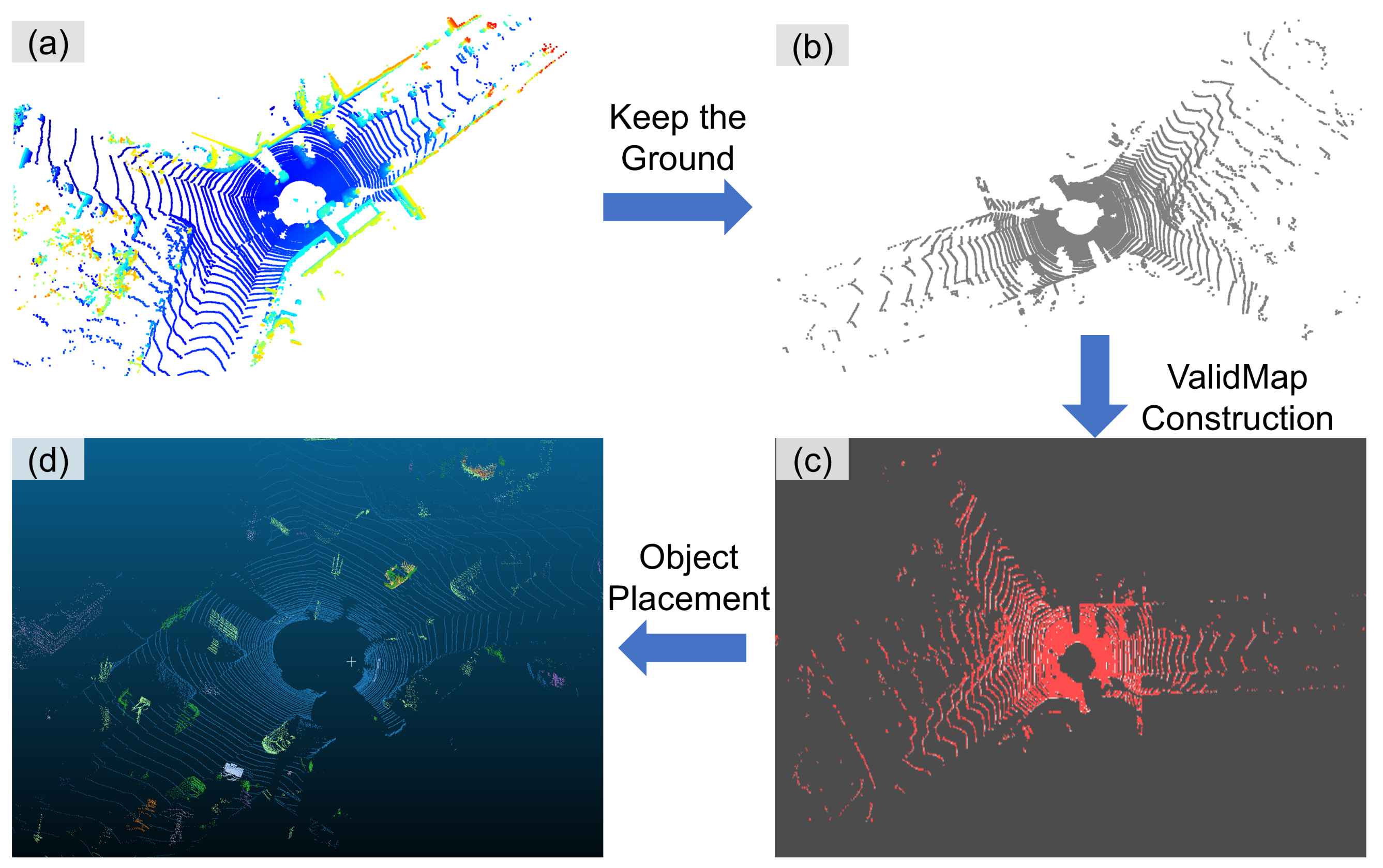} 
    \caption{
        Illustration of the point cloud sequence synthesis process.
        (a) Input LiDAR point cloud. (b) Ground points extracted and retained. (c) ValidMap construction, where valid placement regions are highlighted in red. (d) Final synthesized scene with objects placed into valid regions, resulting in augmented LiDAR point clouds.
    }
    \label{fig:data_augmentation}
    \vspace{-0.1cm}
\end{figure}

% TODO: 首先需要说明一下我们合成 point cloud sequence 的目的, 增加训练数据的多样性?
%In this section, we generate physically plausible synthetic point cloud sequences in an unsupervised manner, which can provide diverse simulated data for training online instance segmentation models.
We propose a novel method for generating synthetic LiDAR point cloud sequences for training purposes, leveraging unsupervised segmentation labels obtained through clustering. Generating synthetic data can significantly enhance training sets by adding diverse scenarios. The method is based on the following key steps:

\textbf{Filter Ground Points.} This step involves separating the LiDAR points that correspond to the ground from those that represent objects. The result is a set of ground points $G_t \subset P_t$ that form the basis for defining the areas where objects can be placed in the scene.

\textbf{ValidMap Construction.}
% Based on the ground segmentation results, we define the ValidMap. Specifically, in the Bird’s Eye View (BEV) plane, any grid cell that contains one or more LiDAR points is considered a valid region. These valid regions are potential areas where objects can be pasted without overlap with other objects or ground points.
Based on the ground segmentation results, we define the ValidMap, which defines the regions of the scene where objects can be placed. The scene is represented in the Bird's Eye View (BEV) plane. Each grid cell in the BEV plane is represented by coordinates $(x_m, y_n)$, where $x_m$ and $y_n$ are the coordinates of the grid cell. A grid cell is considered valid if it contains one or more ground LiDAR points. Mathematically, for each grid cell $(x_m, y_n)$, the ValidMap is defined as a binary map $V$, where:

\begin{equation}\label{eq:valid_map_definition}
\scalebox{0.9}{
	$V(x_m, y_n) =
	\begin{cases} 
		1 & \text{if at least one ground point} \text{ in grid} (x_m, y_n), \\
		0 & \text{otherwise}.
	\end{cases}
	$
}
\end{equation}

In other words, $V(x_m, y_n) = 1$ indicates that the grid cell is valid for placing objects, and $V(x_m, y_n) = 0$ indicates that the grid cell is either occupied by the ground or is invalid due to other reasons.

%\textbf{Object Placement with Collision Detection.}

\textbf{Object Placement with Collision Detection.} With the ValidMap $V$ constructed, we now proceed to place objects into the valid regions. First, we prepare a database $D=\{O_1, O_2, \dots, O_{N_D}\}$ that stores the temporal point clouds of the objects. The $j^{\text{th}}$ object is represented by its point cloud $O_j = \{(x_i^j, y_i^j, z_i^j)\}_{i=1}^{N_j}$, where $N_j$ is the number of points in the point cloud for object $j$. 

When we synthesize a new point cloud sequence, we randomly sample $N_s$ objects from the database $D$, and check the validity of objects for placement. For each object $O_j$, we calculate the 2D bounding box in the Bird’s Eye View (BEV) plane, which encapsulates all the points of  $O_j$. Let the bounding box for object $j$ be represented as $B_j = [x_{\min}^j, x_{\max}^j, y_{\min}^j, y_{\max}^j]$, which defines the minimum and maximum extents of the bounding box along the $x$ and $y$-axes. If the center of box $B_j$ does not lie within the valid region defined by ValidMap $V$, the corresponding object is considered not suitable for current scene.

For collision detection, we check if the bounding box $B_j$ of the new object overlaps with any of the bounding boxes $\{ B_1, B_2, \dots, B_M \}$, where $M$ is the number of objects already placed in the scene. Specifically, we ensure that the bounding box $B_j$ of object $j$ does not overlap with any of the existing bounding boxes.

If the object passes both the valid region check and collision detection, the point cloud $O_j$ is pasted into the corresponding grid cells in the scene. This ensures that the object’s bounding box does not overlap with any other objects or invalid regions, maintaining a physically realistic spatial arrangement in the synthetic scene.

%For collision detection, we check whether the bounding box $B_j$ of the new object overlaps with the bounding boxes of previously placed objects in the scene. Specifically, we ensure that the bounding box $B_j$ of object $j$ does not overlap with any of the bounding boxes $\{B_1, B_2, \dots, B_M\}$, where $M$ is the number of objects already placed in the scene.

%If the object is located at valid region and passes the collision detection xxx.
% The point cloud $O_j$ is then pasted into the corresponding grid cells in the scene, ensuring that its bounding box does not overlap with any other objects or invalid regions, maintaining a realistic spatial arrangement.

To formalize the process of object placement and collision detection, we present the following Algorithm~\ref{object_placement} that outlines the detailed steps for placing objects into valid regions while ensuring spatial consistency in the generated point cloud sequence.

\begin{algorithm}
    \nolinenumbers
    \caption{Object Placement with Collision Detection}\label{object_placement}
    \begin{algorithmic}[1]
        \State \textbf{Input:} ValidMap \( V \), point cloud database \( D = \{ O_1, O_2, \dots, O_{N_D} \} \)
        \For{each object \( j \) sampled from \( D \) (total \( N_s \) objects)}
        \State Calculate the 2D bounding box for object \( j \) in the BEV plane: \( B_j = [x_{\min}^j, x_{\max}^j, y_{\min}^j, y_{\max}^j] \)
        \If{center of \( B_j \) is not within the valid region defined by \( V \)}
        \State \textbf{Reject object \( j \)} \Comment{Object is not suitable for placement}
        \State \textbf{Continue to next object}
        \EndIf
        \For{each previously placed object \( l \) in the scene}
        \If{bounding box \( B_j \) overlaps with \( B_l \)}
        \State \textbf{Reject object \( j \)} \Comment{Object collides with an existing object}
        \State \textbf{Continue to next object}
        \EndIf
        \EndFor
        \If{no overlap with any existing objects}
    %		\State Place object \( j \) in the valid region of the scene according to \( V \)
        \State Paste point cloud \( O_j \) into the scene
        \EndIf
        \EndFor
        \State \textbf{Output:} Updated point cloud sequence with placed objects
    \end{algorithmic}
    \vspace{-0.1cm}
\end{algorithm}

%\begin{algorithm}
%	\caption{Object Placement with Collision Detection}
%	\begin{algorithmic}[1]
%		\State \textbf{Input:} ValidMap \( V \), database of object point clouds \( O_j \)
%			\State 123
%		\For{each object \( j \)}
%		
%		\If{bounding box \( B_j \) overlaps with \( B_i \)}
%			\State \textbf{Reject object \( j \)}
%			\State \textbf{Continue to next object}
%		\EndIf
%		
%		\State Calculate the 2D bounding box for object \( j \): \( B_j = [x_{\min}^j, x_{\max}^j, y_{\min}^j, y_{\max}^j] \)
%		\For{each previously placed object \( i \) in the scene}
%		\If{bounding box \( B_j \) overlaps with \( B_i \)}
%		\State \textbf{Reject object \( j \)}
%		\State \textbf{Continue to next object}
%		\EndIf
%		\EndFor
%		\If{no overlap with any existing objects}
%		\State Place object \( j \) in the valid region of the scene according to \( V \)
%		\State Paste point cloud \( O_j \) into the corresponding grid cells in \( V \)
%		\EndIf
%		\EndFor
%		\State \textbf{Output:} Updated point cloud sequence with placed objects
%	\end{algorithmic}
%\end{algorithm}

\subsection{Flexible Temporal Sampling}

In the original UNIT~\cite{sautier2024unit} method, the Online Instance Segmentation Network primarily relied on adjacent frame pairs $(P_t, P_{t+1})$ for temporal training, where $P_t$ represents the point cloud at time $t$. However, to enhance the model’s ability to capture broader temporal dependencies and improve its generalization capabilities, we improve the sample selection for the temporal training process.

\textbf{Non-adjacent Frame Selection.} Instead of selecting only consecutive frames $(P_t, P_{t+1})$, we introduce a more flexible frame sampling mechanism. Specifically, we randomly select frames that are not immediately adjacent, i.e., $P_t$ and $P_{t+k}$ where $k$ can vary, introducing a broader temporal context for each object instance. This helps the model learn long-term dependencies across time intervals of varying lengths. This selection strategy exposes the model to a wider range of dynamic changes in the scene, capturing more complex temporal relationships.

\textbf{Reversing the Temporal Order.} 
%In traditional temporal models, training usually involves learning from consecutive frames, assuming the temporal continuity of the data. 
%To enable the model to learn from both forward and backward temporal contexts, we reverse the temporal order of point clouds for training. 
Specifically, for each training pair $(P_t, P_{t+k})$, we also sample $(P_{t+k}, P_t)$ as a valid training pair, where $k$ is the time difference between frames. This reversal introduces bidirectional learning, enabling the model to learn from both forward and backward temporal sequences, which enhances its robustness to temporal shifts and improves its ability to capture bidirectional dependencies in motion patterns.

%By combining **Reversing the Temporal Order** and **Non-adjacent Frame Selection**, we create a **flexible temporal training strategy** that enriches the model’s learning experience. The two techniques allow the model to not only understand the forward and backward movement of objects but also learn from long-range temporal dependencies. The training pairs $(P_t, P_{t+k})$ and their reversed counterparts $(P_{t+k}, P_t)$ effectively enhance the temporal context and provide a richer sequence for learning object instance segmentation.

%This combination results in improved generalization across varying temporal intervals and helps the model better adapt to dynamic scenes. The expanded temporal context enables the model to better capture object trajectories, long-term motion patterns, and dynamic variations that would otherwise be overlooked when only considering adjacent frames.

\subsection{Training Objective}
%\textbf{Lable Assignment.}
%\textbf{Instance Mask Loss.} \todo{Given that there is no ordering to the set of instances in a scene and the set of predicted instances, we need to establish correspondences between the two sets during training. To that end, we use bipartite graph matching.}
%We construct a cost matrix $C\in \mathbb{R}^{N_q\times N_o}$
%\todo{We denote the number of queries and pseudo ground-truth instances in a scene by $N_q$ and $N_o$, respectively.}
%The matching cost for a predicted instance $j$ and a target instance $o$ is given by:
%\begin{equation}
%	C(j,o)=\lambda_{\text{dice}}\mathcal{L}_{\text{dice}}(j,o) + \lambda_{\text{BCE}}\mathcal{L}_{\text{BCE}}(j,o)
%\end{equation}
%where $\mathcal{L}_{\text{dice}}$ is the Dice loss and the $\mathcal{L}_{\text{BCE}}$ is the binary cross-entropy loss. 
%\todo{The actual matching between queries and object instances is obtained by ka Hungarian matching with $C$ as the cost matrix.}
%\todo{Once we obtain the certain matching relationship, we can optimize the predicted mask as follows:}
%\begin{equation}
%	\mathcal{L}_{\text{mask}} = \sum_{o=1}^{min(N_q,N_o)} \lambda_{\text{dice}}\mathcal{L}_{\text{dice}}(M(o),o) + \lambda_{\text{BCE}}\mathcal{L}_{\text{BCE}}(M(o),o),
%\end{equation}
%\todo{where $M(o)$ is the index of the matched query of object $o$.} The overall loss is a sum of corresponding loss computations at all layers, as in Mask3D~\cite{schult2022mask3d}.

\textbf{Instance Mask Loss.} Since there is no inherent ordering between the predicted instances and the set of actual instances in a scene, it is necessary to establish correspondences between the two sets during training. To achieve this, we utilize bipartite graph matching.

We construct a cost matrix $C \in \mathbb{R}^{N_q \times N_o}$, where $N_q$ denotes the number of queries and $N_o$ represents the number of pseudo ground-truth instances in the scene. 
The similarity score $S_{ij}$ denotes the dot product between the embedding of $j$-th query and the features of $i$-th point, followed by a sigmoid function. 
The matching cost between a predicted instance $j$ and a target instance $o$ is computed as:
\begin{equation}
C(j,o) = \lambda_{\text{dice}} \mathcal{L}_{\text{dice}}(j,o) + \lambda_{\text{BCE}} \mathcal{L}_{\text{BCE}}(j,o),
\end{equation}
\begin{equation}
	\mathcal{L}_{\text{dice}}(j,o) =  \frac{
	2\textstyle \sum_{i=1}^{N_p} S_{ij}M_{io}
	}{
		\textstyle \sum_{i=1}^{N_p} S_{ij}^2 + \textstyle \sum_{i=1}^{N_p} M_{io}^2
	},
\end{equation}
\begin{equation}
	\mathcal{L}_{\text{BCE}}(j,o) = \sum_{i=1}^{N_p} M_{io}\text{log}(S_{ij}) + (1-M_{io})\text{log}(1-S_{ij})
\end{equation}
where $\mathcal{L}_{\text{dice}}$ is the Dice loss, $\mathcal{L}_{\text{BCE}}$ is the binary cross-entropy loss, $N_p$ is the number of points. The matrix $M \in [0, 1]^{N_p \times N_o}$ serves as a soft assignment indicator between points and object instances, derived from the pseudo-labels. Specifically, $M_{io} = 1$ if point $i$ is associated with object $o$ according to the clustering results, and $M_{io} = 0$ otherwise.

The actual matching between predicted queries and target object instances is performed using the Hungarian algorithm, with the cost matrix $C$ providing the necessary matching costs.

After determining the matching, we optimize the predicted masks by minimizing the following loss:
\begin{equation}
\mathcal{L}_{\text{mask}} = \sum_{o=1}^{\min(N_q, N_o)} \lambda_{\text{dice}} \mathcal{L}_{\text{dice}}(G(o), o) + \lambda_{\text{BCE}} \mathcal{L}_{\text{BCE}}(G(o), o),
\end{equation}
where $G(o)$ is the index of the matched query corresponding to object $o$. This loss function is accumulated across all layers, similar to the approach in Mask3D~\cite{schult2022mask3d}.

\textbf{Time Consistency Loss.} 
Here, we revisit the time consistency loss proposed by UNIT~\cite{sautier2024unit}, which 
enables long-term segmentation and tracking of object instances beyond the training window,
% While the network is trained on two consecutive scans, we aim to ensure temporal coherence over extended periods.

Let an object instance $o$ be segmented at time $t$ and $t+k$ using our pseudo-labeling strategy. We compute the average similarity score (dot product between query embeddings and point features) for the points belonging to $o$ at both $t$ and $t+k$. These scores are then transformed into probability distributions using a softmax function, yielding $H_o^t \in [0, 1]^{N_q}$ and $H_o^{t+k} \in [0, 1]^{N_q}$. To encourage temporal continuity, we minimize the Kullback-Leibler divergence (KL-divergence) between the two distributions:
\begin{equation}
	\mathcal{L}^{\text{cons}}_o = D_{\text{KL}}(H_o^t || H_o^{t+k}).
\end{equation}
We treat $H_o^t$ as the target distribution, and do not backpropagate through it (using a stop-gradient operation), but backpropagation is performed through both time steps, thanks to the auto-regressive architecture.
%In practice, this KL-divergence can be approximated by a cross-entropy loss between $H_o^{t+1}$ and $H_o^t$, while maintaining the stop-gradient on $H_o^t$.

\subsection{Dynamic Weighting Loss for Model Training}
% Cost-sensitive Loss / Dynamic Weighting for Temporal Model Training
% Dynamic sample weighting loss
\textbf{Confidence-based Loss Scaling.}
For the training process, the generated pseudo-labels may contain noisy instances. To alleviate the impact of these noisy instances and focus on more confident predictions, we propose a new loss that dynamically adjusts the contribution of each training pair based on the model's confidence.

%Confidence-based Loss Scaling
To reduce the influence of low-confidence instances, we apply a scaling factor to the loss function based on the model’s confidence for an instance object:
\begin{equation}
	w_{ij}=\max(\epsilon, 1 - \alpha (1 - \mathrm{sg}(S_{ij})))
\end{equation} 
where $\alpha$ is a scaling parameter that controls how much to down-weight low-confidence samples, $\epsilon$ is a small constant value used to ensure numerical stability and avoid the scaling factor $w_{ij}$ becoming too small (or zero) when the model is uncertain about a prediction, $\mathrm{sg}(\cdot)$ denotes the stop-gradient operation.
% This ensures that the gradient flows only through the xxx while treating the xxx as a fixed target during backpropagation.

The scaled instance mask losses can be adjusted as:
%	$$L_{\text{scaled}} = L \times \left( 1 - \alpha \cdot (1 - p) \right)$$
\begin{equation}
	\mathcal{L}_{\text{dice}}^{\text{scaled}}(j,o) =  \frac{
		\textstyle \sum_{i=1}^{N_p} 2w_{ij}S_{ij}M_{io}
	}{
		\textstyle \sum_{i=1}^{N_p} S_{ij}^2 + \textstyle \sum_{i=1}^{N_p} M_{io}^2
	},
\end{equation}
\begin{equation}
	\mathcal{L}_{\text{BCE}}^{\text{scaled}}(j,o) = \sum_{i=1}^{N_p} w_{ij}M_{io}\text{log}(S_{ij}) + (1-M_{io})\text{log}(1-S_{ij}).
\end{equation}

\textbf{Motion-based Weighting for Dynamic Objects.}
In addition to managing noisy training samples, we also address the challenge of uninformative pseudo instances, such as those containing only background or stationary targets. These patches typically correspond to regions with little or no motion (e.g., wall, trees, or grass), which do not contribute significantly to learning dynamic behavior.

To better emphasize dynamic objects—which are more informative and have greater motion between frames—we assign a motion weight vector $\mathbf{A}_{\text{motion}}$ to each instance. The weight is calculated based on the motion of the instance, specifically by computing the displacement of the instance’s centroid between frames. This can be represented as:
\begin{equation}
\mathbf{A}_o^{\text{motion}} = \left\| \mathbf{C}_{t+k}^o - \mathbf{C}_t^o \right\|_2^2,
\end{equation}
where $\mathbf{C}_{t+k}^o$ and $\mathbf{C}_t^o$ represent the centroids of the instance $o$ in frames $P_{t+k}$ and $P_t$, respectively. If an object only appears in one of the two frames, we set its motion weight to zero.

The larger the displacement, the more significant the motion of the object between frames, and thus the higher the weight assigned to that training sample. 
%This ensures that the model focuses more on training with dynamic instances that undergo substantial motion, as they contribute more meaningfully to the task at hand.
To integrate this weighting into training in a stable manner, we normalize the motion weights across all instances using:
\begin{equation}
%\mathbf{A}_{\text{norm}}^i = \frac{\mathbf{A}_{\text{drop}}^i \cdot \mathbf{A}_{\text{motion}}^i}{\sum_{i=1}^{n} \mathbf{A}_{\text{drop}}^i \cdot \mathbf{A}_{\text{motion}}^i}
\mathbf{A}_o^{\text{norm}}  = \frac{ \mathbf{A}_o^{\text{motion}} + \beta}{\sum_{o=1}^{N_o} (\mathbf{A}_o^{ \text{motion}} + \beta)},
\end{equation}
where $\beta$ denotes a small positive constant.
%serves two purposes: (1) it prevents division by zero in cases where all motion weights are zero, and (2) it smooths the normalization, avoiding overly large weights for a few highly dynamic instances. This ensures that all objects receive at least a minimal weight while still favoring those that move.
The final unsupervised loss is then computed as:
%We then train on two consecutive scans by adding the time-consistency loss between the first and second scans:
% \begin{equation}
% 	\resizebox{\hsize}{!}{
% 		$\frac{1}{N} \sum_{o=1}^{N} \mathbf{A}_o^{\text{norm}} \left[ \lambda_{\text{dice}} \, \mathcal{L}_{\text{dice}}^{\text{scaled}}(G(o),o) + \lambda_{\text{BCE}} \, \mathcal{L}_{\text{BCE}}^{\text{scaled}}(G(o),o) + \lambda_\text{cons} \mathcal{L}^\text{cons}_o \right] \>, $
% 	}
% \end{equation}
\begin{align}
\frac{1}{N} \sum_{o=1}^{N} \mathbf{A}_o^{\text{norm}} \big[ \, 
& \lambda_{\text{dice}} \, \mathcal{L}_{\text{dice}}^{\text{scaled}}(G(o), o)  + \lambda_{\text{BCE}} \, \mathcal{L}_{\text{BCE}}^{\text{scaled}}(G(o), o) \nonumber \\
& + \lambda_{\text{cons}} \, \mathcal{L}^\text{cons}_o \, \big] \> .
\end{align}
with $N = \min (N_q, N_o)$.

This dynamic weighting approach enables the model to prioritize informative and confident samples, while de-emphasizing less informative or noisy ones. By combining confidence-based loss scaling and motion-based weighting, we ensure that the model learns effectively from dynamic, informative instances, resulting in better performance and convergence during temporal training.

\section{Experiments}\label{sec:experiments}
\subsection{Implementation Details}
%We test our method on widely used three benchmarks: SemanticKITTI~\cite{behley2019semantickitti} PandaSet~\cite{xiao2021pandaset}
%3) nuScenes~\cite{caesar2020nuscenes}
%More details about these datasets, please see the supplementary materials.
We evaluate our method on three widely used benchmarks: SemanticKITTI\cite{behley2019semantickitti}, PandaSet\cite{xiao2021pandaset}, and nuScenes~\cite{caesar2020nuscenes}. Additional details about these datasets, as well as the implementation details of our method, are provided in the supplementary material.

\subsection{Baselines}
3DUIS~\cite{3duis} is a scan-level unsupervised instance segmentation method based on graph-cut optimization. It segments each LiDAR frame independently, without leveraging temporal information. TARL-Seg~\cite{tarl} and 4D-Seg~\cite{sautier2024unit} both produce spatio-temporal segments by clustering point features over a short window of registered scans using HDBSCAN. While TARL-Seg reuses the segmentation component from the TARL framework, 4D-Seg further applies a voxel grid sampling and a long time window for clustering.

To enable long-term tracking, we extend all three baselines with a lightweight temporal linking strategy. After generating 4D segments within individual time windows, we connect instances across windows by aligning the last scan of one window with the first of the next (using ground-truth ego poses), and matching object instances via convex hull IoU. When two instances overlap above a threshold (0.5 IoU), we assign a consistent ID. We refer to the improved versions as 3DUIS++, TARL-seg++, and 4D-Seg++.

UNIT~\cite{sautier2024unit} is a fully unsupervised and online 3D instance segmentation method. It processes LiDAR scans sequentially, associating points to object tracks over time. UNIT maintains temporal consistency to segment and track instances in real-time, making it well-suited for online applications.

%\subsection{Implementation Details}

\subsection{Metrics}
\textbf{Association Score.} The primary evaluation metric we use is the association score, originally defined in~\cite{aygun20214d} for 4D panoptic segmentation of LiDAR data and applied in~\cite{3duis,sautier2024unit} for unsupervised, class-agnostic instance segmentation. We provide the definition here for completeness.

Let $\mathcal{G}$ represent the set of manually annotated object instances (sets of points) used as ground truth for evaluation, which are not part of the training data, and let $\mathcal{S}$ be the set of predicted segments (objects) from our model. Each object in both $\mathcal{G}$ and $\mathcal{S}$ corresponds to a 4D segment, containing the list of points associated with that object over time. The temporal association score $\mathrm{S_{assoc}^{temp}}$ is given by:
\vspace{-0.1cm}
\begin{equation}
	\mathrm{S_{assoc}^{temp}} = \dfrac{1}{|\mathcal{G}|} \sum\limits_{g \in \mathcal{G}}\dfrac{1}{|g|} \sum\limits_{\substack{s \in S \\ \ s \cap g \neq 0}} \mathrm{TPA}(s,g) \mathrm{IoU}(s,g),
\end{equation}
\vspace{-0.1cm}
where $\mathrm{TPA}(s,g) = |s \cap g|$ represents the number of true positive associations between the segments $s$ and $g$, and $\mathrm{IoU}$ stands for the intersection-over-union.

For cases where temporal consistency is not considered, a non-temporal association score $\mathrm{S_{assoc}}$ can be computed by treating the objects and predictions in separate scans as distinct instances. 
%This association metric, denoted as $\mathrm{S\_{assoc}}$, is used in~\cite{3duis}, as their method operates in a scan-wise fashion.

\textbf{Best IoU.} In addition to the association score, we also use a complementary metric called Best IoU~\cite{sautier2024unit}. For each object in $\mathcal{G}$, we identify the most overlapping segments in $\mathcal{S}$ based on the highest IoU, and then compute the average IoU over all objects:
\vspace{-0.1cm}
\begin{equation}
	\mathrm{IoU^*} = \dfrac{1}{|\mathcal{G}|} \sum\limits_{g \in \mathcal{G}} \max\limits_{s \in S} \mathrm{IoU}(s,g).
\end{equation}
\vspace{-0.1cm}
This metric evaluates 4D segments, and a higher score indicates better temporal consistency. Notably, a predicted segment $s$ can match multiple ground-truth objects $g$ and $g'$ ($g \neq g'$), making this metric a useful measure of the best IoU achievable with the available predictions.

%Limitations. The metrics defined above require 4D manual annotations for each object instance. However, in most existing datasets, such annotations are only available for thing categories (e.g., cars, pedestrians). As a result, while our method can segment and track both thing and stuff objects, the evaluation metrics are limited to assessing performance only on a subset of object types.

% \newcommand*\rotext{\multicolumn{1}{R{90}{1em}}}
\setlength{\tabcolsep}{5pt}
\begin{table}[t]
	\centering
	\caption{Unsupervised 3D instance segmentation results on the SemanticKITTI dataset.
		%		\other{All scores are computed on the validation set of SemanticKITTI. The association scores are computed using the code of \cite{aygun20214d}, which, by default, is only applied on segments of more than $50$ points for any given scan; we report the corresponding scores with and without filtering.}
	}
	\label{table:results_sk}
	\begin{tabular}{c|c|ccc|cc} 
		\toprule
		\multirow{2}{*}{Method} & \multirow{2}{*}{Online} & \multicolumn{3}{c|}{Unfiltered}   & \multicolumn{2}{c}{Filtered}      \\
		& & $\mathrm{S_{assoc}^{temp}}$  &  $\mathrm{IoU^*}$ & $\mathrm{S_{assoc}}$ & $\mathrm{S_{assoc}^{temp}}$  & $\mathrm{S_{assoc}}$   \\
		\midrule
		TARL-Seg~\cite{tarl}  & $\times$ & 0.231    & 0.353  & 0.668      & 0.264    & 0.735       \\
		TARL-Seg++~\cite{tarl} & $\times$ & 0.317    & 0.446  & 0.617      & 0.370    & 0.678       \\
		4D-Seg~\cite{sautier2024unit}     & $\times$ & 0.421    & 0.529  & 0.667      & 0.486    & 0.784       \\
		4D-Seg++~\cite{sautier2024unit}   & $\times$ & 0.447    & 0.513  & 0.647      & 0.512    & 0.762       \\
		3DUIS++~\cite{3duis}    & $\times$ & 0.116    & 0.214  & 0.550      & 0.148    & 0.769       \\
		\midrule
		UNIT~\cite{sautier2024unit}       & $\checkmark$ & 0.482    & 0.568  & 0.696      & 0.563    & 0.790       \\
		Ours       & $\checkmark$ & \bf 0.523 & \bf 0.602  & \bf 0.725   & \bf 0.601 & \bf 0.818       \\
		\bottomrule
	\end{tabular}
    \vspace{-0.1cm}
\end{table}
\setlength{\tabcolsep}{6pt}

\textbf{Filtered Metrics.} Following the methodology of 3DUIS~\cite{3duis} and \cite{aygun20214d}, the metrics often include a filter that excludes all 3D ground-truth segments with fewer than 50 points. This ``filtered" version of the metrics is provided for comparison purposes. However, we should note that this filtering removes smaller or more distant objects from the evaluation. To offer a more comprehensive performance assessment, we also compute the metrics without filtering, as removing these objects could unnecessarily simplify the task and doesn't align with the real-world scenario, especially in temporal settings where the point count per object can vary across scans.

\setlength{\tabcolsep}{5pt}
\begin{table}[t]
	\centering
	\caption{Unsupervised 3D instance segmentation results on nuScenes validation dataset.}
    \vspace{-0.1cm}
	\label{table:results_ns}
	\begin{tabular}{c|c|ccc|cc}
		\toprule
		\multirow{2}{*}{Method} & \multirow{2}{*}{Online} & \multicolumn{3}{c|}{Unfiltered}  & \multicolumn{2}{c}{Filtered}      \\
		& & $\mathrm{S_{assoc}^{temp}}$  &  $\mathrm{IoU^*}$ & $\mathrm{S_{assoc}}$ & $\mathrm{S_{assoc}^{temp}}$  & $\mathrm{S_{assoc}}$   \\
		\midrule
		TARL-Seg~\cite{tarl}  & $\times$ & 0.085 & 0.156 & 0.189 & 0.253 & 0.475       \\
		4D-Seg~\cite{sautier2024unit}     & $\times$ & \bf0.163 & 0.246 & 0.287 & 0.350 & 0.546     \\
		\midrule
		UNIT~\cite{sautier2024unit}        & $\checkmark$ & 0.126	& 0.221	& 0.356	& 0.270	& 0.561 \\
		Ours        & $\checkmark$ & 0.162	& \bf0.262	& \bf0.405	& \bf0.354	& \bf0.575 \\
		\bottomrule
	\end{tabular}
    \vspace{-0.1cm}
\end{table}
\setlength{\tabcolsep}{6pt}

\begin{figure*}[t]
    \centering
    \includegraphics[width=0.9\textwidth]{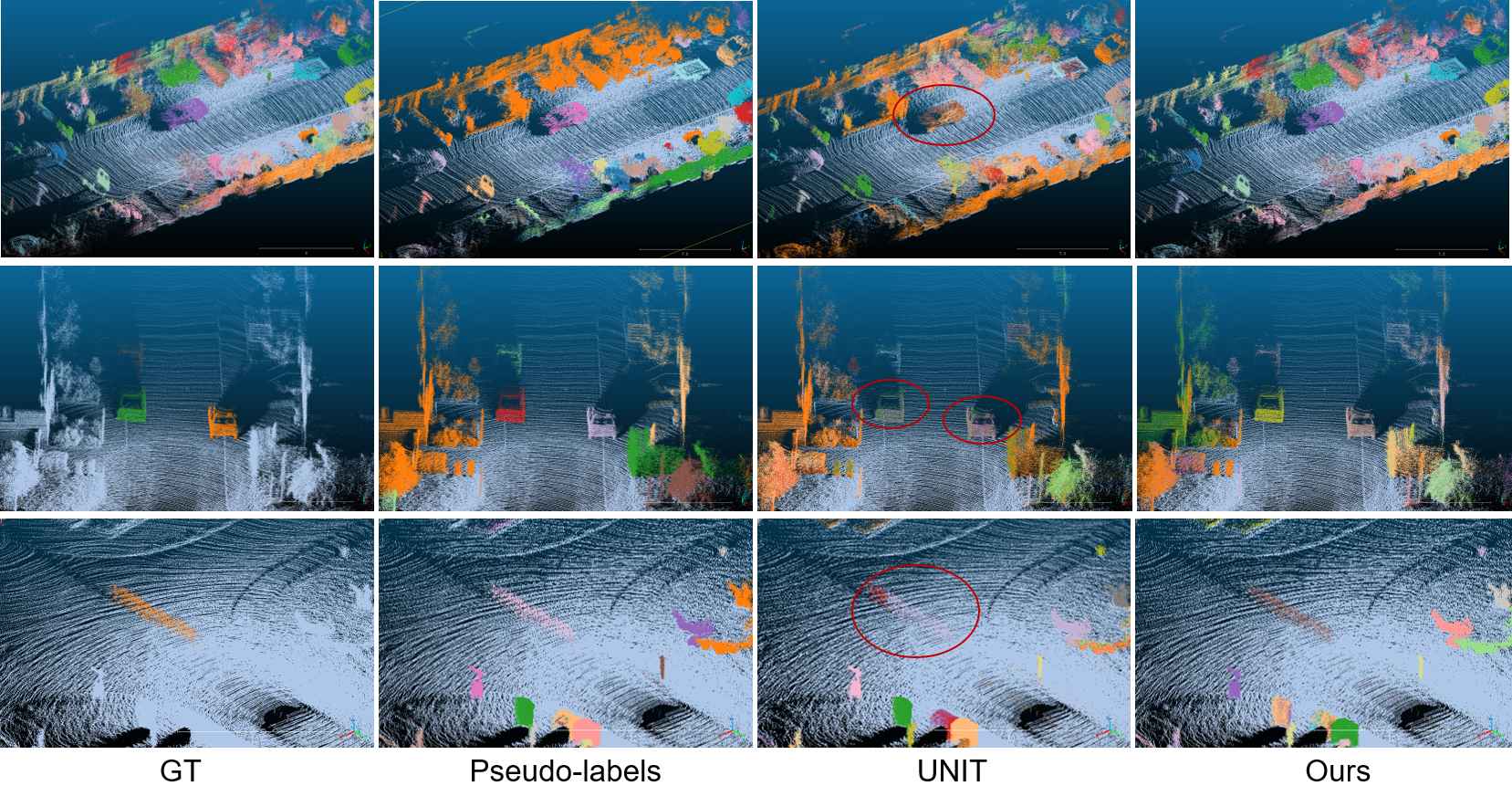} 
    \caption{
        Visual results on SemanticKITTI. From left to right: ground-truth labels (GT), pseudo-labels, predictions from UNIT, and results from ours. Our approach produces cleaner and more consistent instance segmentation, especially in challenging regions (errors from UNIT are highlighted with red circles). All results are class-agnostic, where colors indicate different instances.
    }
    \label{fig:vis_sk}
    \vspace{-0.1cm}
\end{figure*}

\subsection{Main Result}
\textbf{Results on SemanticKITTI.} On the SemanticKITTI validation set, we compare our method against several state-of-the-art unsupervised 3D instance segmentation approaches (see Table~\ref{table:results_sk}).
In the unfiltered case, our method achieves a temporal association score of 0.523, an $\mathrm{IoU^*}$ of 0.602, and an association score of 0.725, which correspond to consistent gains over UNIT of +0.041, +0.034, and +0.029, respectively.
%The results show that offline methods such as TARL-Seg, 4D-Seg, and 3DUIS++ struggle to capture consistent temporal information and generally perform worse than online approaches. 
Similarly, under the filtered setting, our approach reaches 0.601 for temporal association and 0.818 for association score, again outperforming UNIT and confirming the robustness of our method.
% Overall, these results demonstrate that our framework not only improves the short-term segmentation quality but also ensures stronger long-term consistency in tracking object instances.
The superior performance across both unfiltered and filtered metrics highlights the effectiveness of our design, which together enable stable and reliable unsupervised instance segmentation in sequential 3D data.
Fig.~\ref{fig:vis_sk} shows representative qualitative results on the SemanticKITTI dataset. Compared with pseudo-labels and UNIT, our method produces cleaner and more consistent instance masks, with clearer object boundaries. 
%As highlighted in the red circles, UNIT often confuses nearby objects or generates incomplete segments, while our approach is able to recover more accurate and temporally coherent instances. These improvements demonstrate the effectiveness of our method in handling complex and dynamic LiDAR scenes.

%\textbf{Results on nuScenes.} 
%For the nuScenes dataset, the point clouds have lower density than the SemanticKITTI and PandaSet datasets.
%UNIT does not surpass the 4D-seg. \other{Sautier et al~\cite{sautier2024unit} hypothesize that the small point density leads to worse pseudolabels for small or distant objects, which in turn creates more noise on the segments used in the consistency loss.}
%1) Our method achieves consistent improvement compared with the baseline UNIT.
%2) Except for the $\mathrm{S_{assoc}^{temp}}$ metric, our method outperforms the TARL-Seg and 4D-Seg. The results demonstrate our proposed method achieves substantial improvement compared with the baselines.

\textbf{Results on nuScenes.} 
On the nuScenes validation set, the LiDAR scans are much sparser than in SemanticKITTI and PandaSet, making segmentation more difficult. As shown in Table~\ref{table:results_ns}, UNIT does not surpass 4D-Seg, likely because the reduced point density leads to poorer pseudo-labels for small or distant objects and increases noise in the consistency loss. Nevertheless, our method consistently improves upon UNIT across all metrics. It outperforms other methods in most metrics, while only falling slightly behind 4D-Seg on the temporal association score. These results demonstrate that our approach remains effective even under the challenging conditions of low-density point clouds.

\setlength{\tabcolsep}{5pt}
\begin{table}[t]
	\centering
	\caption{Unsupervised 3D instance segmentation results on PandaSet-GT validation dataset.}
	\label{table:results_pandaset}
	\begin{tabular}{c|c|ccc|cc}
		\toprule
		\multirow{2}{*}{Method} & \multirow{2}{*}{Online} & \multicolumn{3}{c|}{Unfiltered}                                                                                          & \multicolumn{2}{c}{Filtered}      \\
		& & $\mathrm{S_{assoc}^{temp}}$  &  $\mathrm{IoU^*}$ & $\mathrm{S_{assoc}}$ & $\mathrm{S_{assoc}^{temp}}$  & $\mathrm{S_{assoc}}$   \\
		\midrule
		TARL-Seg~\cite{tarl}   & $\times$ & 0.206 & 0.286 & 0.369 & 0.390 & \bf0.757  \\
		4D-Seg~\cite{sautier2024unit}     & $\times$ & \bf0.332 & 0.399 & 0.492 & 0.503 & 0.740  \\
		\midrule
		UNIT~\cite{sautier2024unit}  & $\checkmark$  & 0.209	& 0.310 &	0.531	& 0.351 &	0.688 \\
		Ours       & $\checkmark$  & 0.285	& 	\bf0.372	& \bf0.579	& \bf0.456	& 0.737   \\
		\bottomrule
	\end{tabular}
\end{table}
\setlength{\tabcolsep}{6pt}

\textbf{Results on PandaSet-GT.} 
%1) Our method also achieves consistent improvement compared with the baseline UNIT on this dataset.
%2) Online methods, UNIT and ours, underperform 4D-seg in some metrics.
%\other{We recall nevertheless that UNIT work online and without accumulated point clouds and registration of scans, unlike 4D-Seg.}
%On the PandaSet-GT validation set, our method also achieves consistent improvements over the baseline UNIT across most metrics. 
We report the results on PandarGT in Table~\ref{table:results_pandaset}.
While both online approaches, UNIT and ours, still fall behind 4D-Seg on certain measures, this gap is largely due to the offline nature of 4D-Seg, which benefits from a much larger temporal window and access to accumulated point clouds with explicit scan registration. In contrast, our method operates fully online, without relying on such global information, yet still attains competitive results and delivers clear gains in both segmentation accuracy and association quality compared with UNIT.

% We compare our method against all considered baselines on the SemanticKITTI dataset, and the results are reported in Table 1. In the single-scan setting, UNIT already outperforms 3DUIS, highlighting the benefit of training an instance segmentation network end-to-end instead of relying on graph-cut based heuristics. In the online setting, we observe that post-processing the segments produced by 3DUIS is not sufficient to reach the performance of TARL-Seg and 4D-Seg on both the temporal association score and $IoU^*$. Between these two methods, 4D-Seg performs better than TARL-Seg, which can be attributed to its longer temporal window, since both approaches achieve similar scores on the scan-wise association metric.

% Our method builds on UNIT by incorporating dynamic weighting and improved temporal consistency, which leads to consistent improvements across all metrics. In particular, our approach surpasses UNIT by a clear margin both in the unfiltered and filtered settings, demonstrating stronger short-term segmentation quality as well as better long-term temporal coherence. The best overall scores are achieved with our method, confirming the effectiveness of our design choices for unsupervised online 3D instance segmentation.

\setlength{\tabcolsep}{10pt}
\begin{table}[t]
	\centering
	\caption{Ablation study of key components. ``PCSS" denotes point cloud sequence synthesis; ``FTS" means flexible temporal sampling; ``DWL" is dynamic weighting loss for model training.}
	\label{table:ablation_components}
	\begin{tabular}{ccc|ccc} 
		\toprule
		PCSS & FTS & DWL & $\mathrm{S_{assoc}^{temp}}$  &  $\mathrm{IoU^*}$ & $\mathrm{S_{assoc}}$     \\
		\midrule
		$\times$  & $\times$ & $\times$ & 0.482 & 0.568 & 0.696  \\
		\midrule
		$\checkmark$ &  $\times$   &  $\times$     & 0.505 & 0.584 & 0.716  \\
		$\times$ &  $\checkmark$   &  $\times$     & 0.490 & 0.574 &	0.701 \\
		$\times$ &  $\times$   &  $\checkmark$     & 0.497 &	0.582 &	0.709 \\
		\midrule
		$\times$ &  $\checkmark$   &  $\checkmark$ & 0.504 &	0.583 &	0.712 \\
		$\checkmark$ &  $\times$   &  $\checkmark$ & 0.515 &	0.595 &	0.719 \\
		$\checkmark$ &  $\checkmark$   &  $\times$ & 0.510 &	0.592 &	0.718 \\
		\midrule
		$\checkmark$ &  $\checkmark$   &  $\checkmark$ & \bf0.523 & \bf0.602 & \bf0.725  \\
		\bottomrule
	\end{tabular}
\end{table}
\setlength{\tabcolsep}{5pt}

\subsection{Ablation Study}
\textbf{Effect of Key Components.}
The results of the ablation study in Table~\ref{table:ablation_components} demonstrate that each component contributes to the overall performance. The baseline model achieves a temporal association score of 0.482, $\mathrm{IoU^*}$ of 0.568, and an association score of 0.696. Introducing point cloud sequence synthesis (PCSS) alone improves performance across all metrics. Adding flexible temporal sampling (FTS) and dynamic weighting loss (DWL) also boosts performance, with PCSS + DWL showing the most significant improvement. Combining all three components yields the best results, with scores of 0.523, 0.602, and 0.725 for temporal association, $\mathrm{IoU^*}$, and association score, respectively.

\setlength{\tabcolsep}{15pt}
\begin{table}[t]
	\centering
	\caption{Ablation study on the impact of different sampling numbers $N_s$ on performance.}
	\label{table:ablation_sample_num}
	\begin{tabular}{c|ccc} 
		\toprule
		$N_s$    & $\mathrm{S_{assoc}^{temp}}$  &  $\mathrm{IoU^*}$ & $\mathrm{S_{assoc}}$     \\
		\midrule
		200  & 0.495 & 0.578 & 0.707  \\
		400  & 0.506 & 0.585 & 0.714  \\
		600  & \bf0.523 & \bf0.602 & \bf0.725  \\
		800  & 0.515 & 0.594 & 0.716  \\
		1000 & 0.507 & 0.593 & 0.715  \\
		\bottomrule
	\end{tabular}
\end{table}
\setlength{\tabcolsep}{5pt}

%\textbf{Hyper-parameters and design choices.}
\textbf{Effect of Different Sampling Numbers.} 
We further study the influence of the sampling number $N_s$ on model performance. As shown in Table~\ref{table:ablation_sample_num}, increasing $N_s$ from 200 to 600 steadily improves all metrics, with the best results obtained at $N_s = 600$ ($\mathrm{S_{assoc}^{temp}} = 0.523$, $\mathrm{IoU^*} = 0.602$, $\mathrm{S_{assoc}} = 0.725$). Beyond this point, performance slightly decreases when using larger sampling numbers such as 800 or 1000, indicating that excessively large sampling may introduce redundancy or noise rather than further benefits. These results suggest that a moderate sampling size provides the best trade-off between diversity and stability for effective training.

\setlength{\tabcolsep}{15pt}
\begin{table}[t]
	\centering
	\caption{\major{Influence of different sampling strategies for training. ``RTO" denotes Reversing the Temporal Order; ``NFS" means Non-adjacent Frame Selection. ``Fixed-gap" denotes non-random fixed-gap sampling where interval is 2.}}
	\label{table:ablation_sampling}
	\begin{tabular}{c|ccc} 
		\toprule
		Method      & $\mathrm{S_{assoc}^{temp}}$  &  $\mathrm{IoU^*}$ & $\mathrm{S_{assoc}}$     \\
		\midrule
		Ours        & \bf0.523 & \bf0.602 & \bf0.725  \\
		Ours w/o RTO & 0.521 & 0.599 & 0.723  \\
		Ours w/o NFS & 0.518 & 0.597 & 0.720  \\
		\major{Fixed-gap}    & 0.508 & 0.588 & 0.711  \\
		\bottomrule
	\end{tabular}
\end{table}
\setlength{\tabcolsep}{5pt}

\textbf{Effect of Different Sampling Strategies.}
\major{Table~\ref{table:ablation_sampling} evaluates the impact of our proposed sampling strategies during training. Removing either reversing the temporal order (RTO) or non-adjacent frame selection (NFS) consistently lowers performance across all metrics. Without RTO, the scores drop slightly to 0.521 in temporal association, 0.599 in $\mathrm{IoU^*}$, and 0.723 in association score. Similarly, excluding NFS reduces the scores further to 0.518, 0.597, and 0.720, respectively. The full model with both strategies enabled achieves the best results, confirming that RTO and NFS are complementary and together improve temporal modeling and overall segmentation quality. And our proposed sampling strategy also outperforms the fixed‑gap sampling across all three metrics.}

\setlength{\tabcolsep}{15pt}
\begin{table}[htp]
	\centering
	\caption{Influence of different DWL components. CLS: Confidence-based Loss Scaling; MWOD: Motion-based Weighting for Dynamic Objects.}
    \vspace{-0.1cm}
	\label{table:ablation_loss}
	\begin{tabular}{cc|ccc}
		\toprule
		CLS & MWDO & $\mathrm{S_{assoc}^{temp}}$  &  $\mathrm{IoU^*}$ & $\mathrm{S_{assoc}}$     \\
		\midrule
		$\times$   & $\times$    & 0.482 & 0.568 & 0.696  \\
		$\checkmark$   & $\times$    & 0.492 & 0.577 & 0.705  \\
		$\times$   & $\checkmark$    & 0.488 & 0.572 & 0.701  \\
		$\checkmark$   & $\checkmark$    & \bf0.497 & \bf0.582 & \bf0.709  \\
		\bottomrule
	\end{tabular}
\end{table}
\setlength{\tabcolsep}{5pt}

\textbf{Effectiveness of Different DWL Components.} 
Table~\ref{table:ablation_loss} shows that both confidence-based loss scaling (CLS) and motion-based weighting for dynamic objects (MWDO) contribute to performance gains. Each component individually improves over the baseline, and their combination achieves the best results (0.497, 0.582, 0.709), confirming that they are complementary in reducing noisy supervision and emphasizing informative instances during training.

%we have conducted additional experiments comparing our synthetic strategy to the widely used Copy-Paste augmentation method.
\textbf{\major{Effectiveness of Point Cloud Sequence Synthesis.}} 
\major{As shown in Table~\ref{table:ablation_syn}, the results highlight that our approach outperforms the Copy-Paste method across all metrics, including $\mathrm{IoU^*}$ and $\mathrm{S_{assoc}}$. We observe the highest performance with a substantial improvement in $\mathrm{IoU^*}$ (0.589 vs. 0.575) and $\mathrm{S_{assoc}}$ (0.716 vs. 0.708).
\vspace{-0.1cm}}

\setlength{\tabcolsep}{10pt}
\begin{table}[t]
	\centering
	\caption{\major{Comparison between the common copy-paste strategy and our point cloud sequence synthesis method.}}
	\label{table:ablation_syn}
	\begin{tabular}{cc|ccc} 
		\toprule
		Copy-Paste  & Ours  & $\mathrm{S_{assoc}^{temp}}$  &  $\mathrm{IoU^*}$ & $\mathrm{S_{assoc}}$     \\
		\midrule
		$\times$   & $\times$ & 0.482   &  0.568   &  0.696  \\
		$\checkmark$   & $\times$& 0.496   &  0.575   &  0.708  \\
		$\times$   & $\checkmark$ & \bf0.505   & \bf 0.589   &  \bf0.716  \\
		\bottomrule
	\end{tabular}
\end{table}
\setlength{\tabcolsep}{5pt}

\setlength{\tabcolsep}{4pt}
\begin{table}[t]
	\centering
	\caption{
		\major{Efficiency and performance comparison on the SemanticKITTI validation split. Model size, inference latency, frames per second (FPS), and $\mathrm{S_{assoc}^{temp}}$ are reported.}
	}
	\begin{tabular}{l|cccccc}
		\hline
		Method & \makecell{Latency\\(s)} & \makecell{FPS} & \makecell{Training\\Time (h)} & \makecell{Memory\\(GB)} & \makecell{Model\\Size(M)} & $\mathrm{S_{assoc}^{temp}}$ \\
		\hline
		TARL-Seg~\cite{tarl} & 0.97 & 1.03 & 0 & 0 & 0 & 0.231 \\
		4D-Seg~\cite{sautier2024unit} & 1.52 & 0.65 & 0 & 0 & 0 & 0.421 \\
		3DUIS++~\cite{3duis} & 74.09 & 0.01 & 189 & 0.98 & 167 & 0.447 \\
		UNIT~\cite{sautier2024unit} & 0.49 & 2.01 & 48 & 4.16 & 909 & 0.482 \\
		Ours & 0.49 & 2.01 & 50 & 4.16 & 909 & 0.523 \\
		\hline
	\end{tabular}
	\label{tab:efficiency_comparison}
\end{table}
\setlength{\tabcolsep}{5pt}

\subsection{Analysis of Efficiency}
\major{Table~\ref{tab:efficiency_comparison} presents a comprehensive comparison of computational efficiency and performance across different methods on the SemanticKITTI validation split. By sharing the same network architecture, our method introduces no additional runtime overhead during inference compared to UNIT, thus maintaining identical inference latency and frames per second (FPS). Notably, although 3DUIS++ achieves a competitive $\mathrm{S_{assoc}^{temp}}$ score (0.447), its computational cost is prohibitively high (latency of 74.09 s, FPS of 0.01, and training time of 189 hours), making it unsuitable for online applications. In contrast, our approach retains UNIT’s real-time capability while improving segmentation accuracy, offering a favorable balance between performance and efficiency for practical deployment.}

%\begin{table}[htp]
%	\centering
%	\caption{Summary of Point Cloud Self-Supervised Learning Methods.}
%	\begin{tabular}{lllllll}
%		\multirow{2}{*}{Method}   & Latency & FPS  & Training Time & Memory (G) & Model (Size) & S\^{}temp  \\
%		TARL-Seg & 0.97    & 1.03 & 0             & 0          & 0            & 0.231      \\
%		4D-Seg   & 1.52    & 0.65 & 0             & 0          & 0            & 0.421      \\
%		3DUIS++  & 74.09   & 0.01 & 189           & 0.98       & 167M         & 0.447      \\
%		UNIT     & 0.49    & 2.01 & 48            & 4.16       & 909M         & 0.482      \\
%		Ours     & 0.49    & 2.01 & 50            & 4.16       & 909M         & 0.523     
%	\end{tabular}
%\end{table}

\subsection{Qualitative Results}
\begin{figure}[t]
    \centering
    \includegraphics[width=0.5\textwidth]{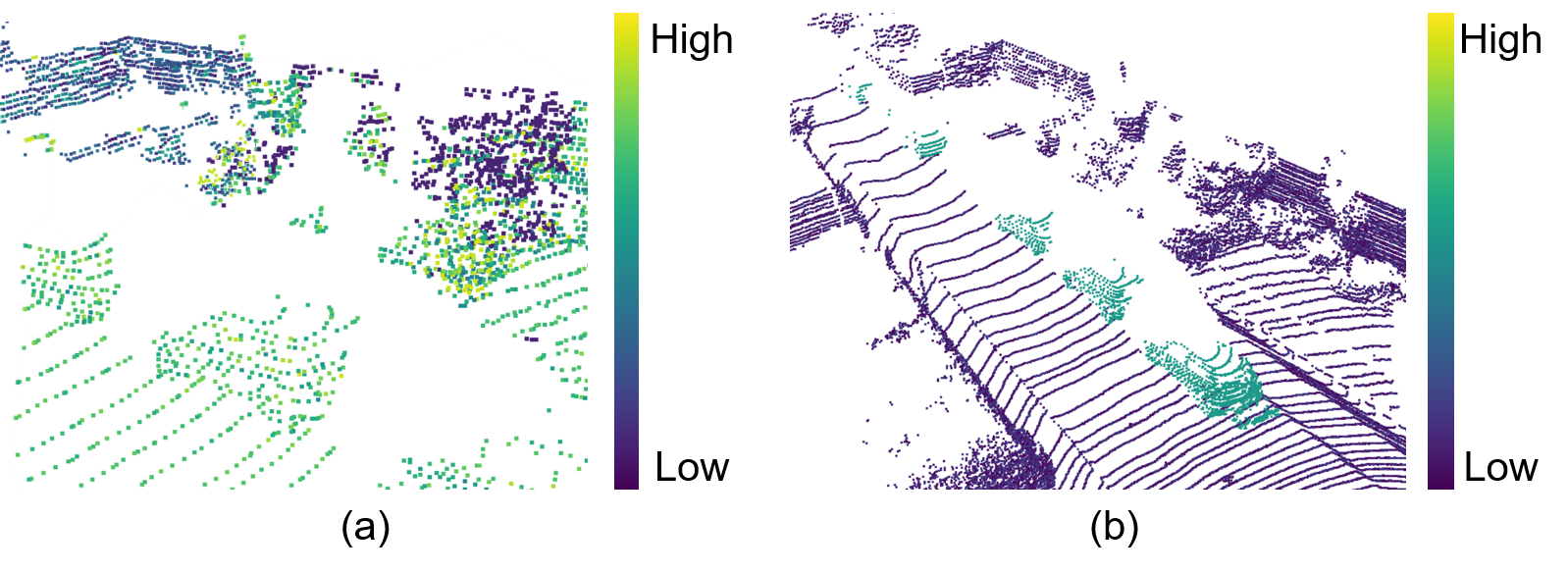} 
    \caption{
        Visualization of the (a) confidence-based scaling factor and (b) motion-based weight vector for training.
    }
    \label{fig:vis_loss}
    \vspace{-0.15cm}
\end{figure}

\begin{figure*}[htbp]
    \centering
    \includegraphics[width=\textwidth]{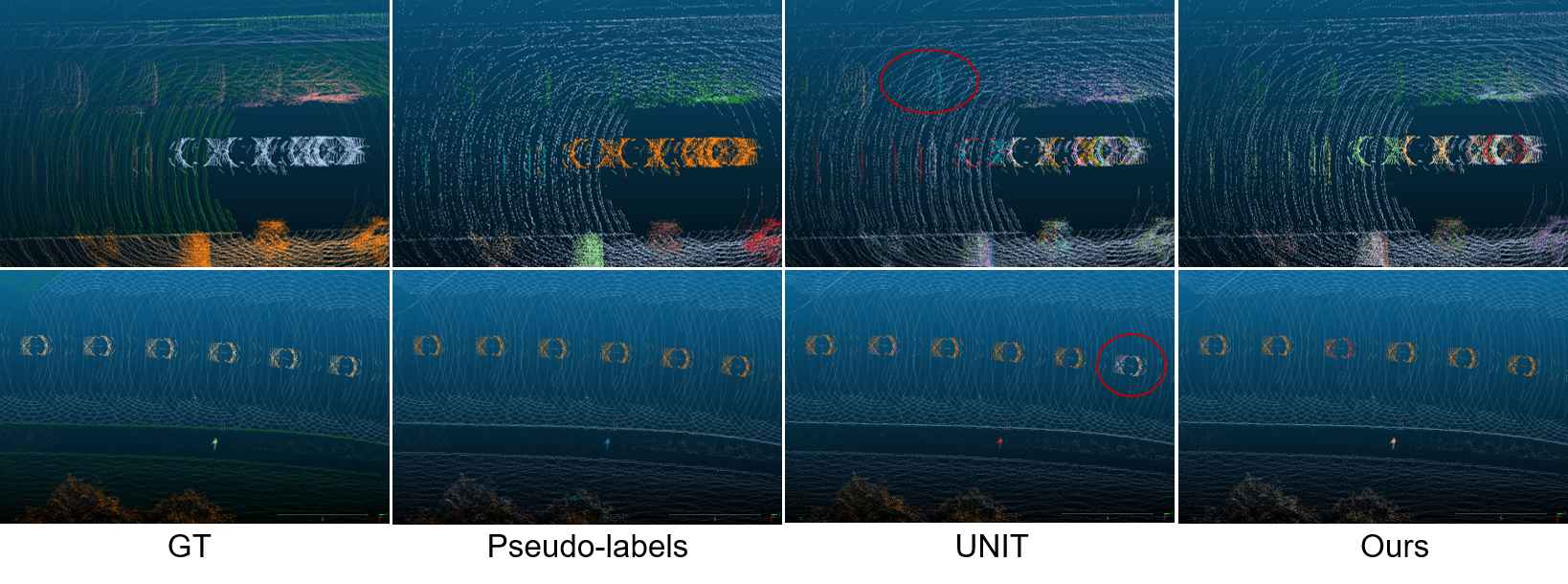}
    \vspace{-0.1cm}
    \caption{
    %		Visual results on nuScenes dataset. Best seen zoomed.
        Visual results on the nuScenes dataset. Our method produces more consistent instance masks compared with the baseline UNIT. Best viewed in zoom.
    }
    \label{fig:vis_ns}
    \vspace{-0.1cm}
\end{figure*}

\begin{figure*}[t]
	\centering
	\includegraphics[width=\textwidth]{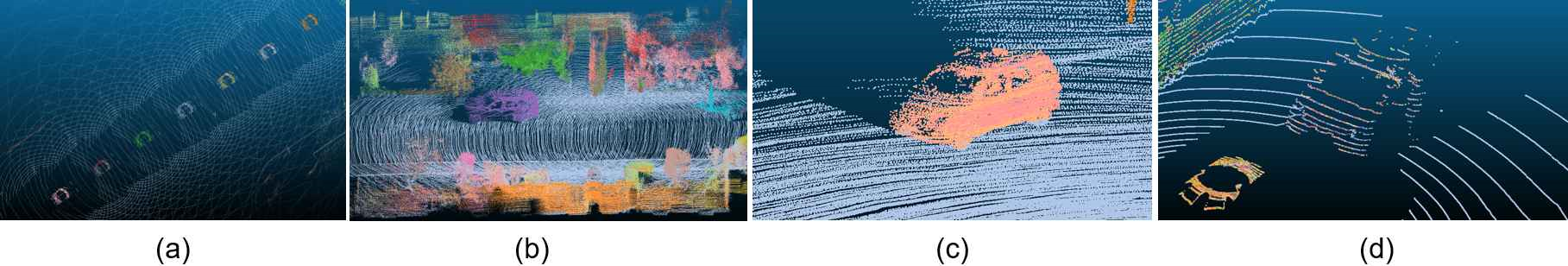}
	%	\vspace{-0.4cm}
	\caption{
		\major{Qualitative examples of failure cases encountered by our method: (a) Tracking failure under fast object motion, (b) Segmentation breakdown for non-structured surfaces (wall), (c, d) A single car instance incorrectly divided into multiple IDs, highlighting limitations in point cloud representation and temporal dependency capture.}
	}
	\label{fig:failures}
	\vspace{-0.15cm}
\end{figure*}

Figure~\ref{fig:vis_loss} illustrates how the proposed loss weighting mechanisms affect training. In Fig.~\ref{fig:vis_loss}\textcolor{red}{(a)}, points that may have been assigned inaccurate instance labels due to spatio-temporal clustering are down-weighted by the confidence-based scaling factor, reducing their influence on the loss. In Fig.~\ref{fig:vis_loss}\textcolor{red}{(b)}, instances that exhibit noticeable displacement between frames are assigned larger weights by the motion-based vector, ensuring that dynamic and informative objects contribute more strongly during training.

Figure~\ref{fig:vis_ns} presents qualitative comparisons on the nuScenes dataset. Our method yields more accurate and temporally consistent instance segmentation compared with pseudo-labels and UNIT. In nuScenes, UNIT introduces inconsistent predictions (highlighted with red circles), whereas our approach produces temporally consistent masks with clearer object boundaries. 
%On PandaSet, which contains complex urban scenes with diverse object types, our method better preserves object shapes and reduces over-segmentation, leading to results that are closer to the ground-truth annotations. These improvements highlight the robustness and generalization of our approach across different datasets.

\major{While the proposed method performs well in many scenarios, we observe some limitations. For instance, when objects move at high relative speeds or are partially occluded, the temporal association mechanism struggles to maintain consistent instance IDs, leading to tracking failures, as shown in Fig.\ref{fig:failures}(a). Additionally, non-structural objects like walls, fences, and vegetation, which lack clear boundaries, are sometimes over-segmented or incorrectly merged into the background, as seen in Fig.\ref{fig:failures}(b). Another issue arises when sparse point clouds or insufficient temporal cues cause a single object, such as a car, to be split into multiple IDs (Fig.~\ref{fig:failures}(c,d)). These challenges highlight the complexity of unsupervised instance segmentation and indicate areas for future improvements.}

\section{Conclusion}\label{sec:conclusion}
In this paper, we presented a novel framework for unsupervised online 3D instance segmentation that addresses key limitations of existing approaches. Our method enriches the training distribution through synthetic point cloud sequence generation, captures both short- and long-range temporal dependencies with flexible sampling, and emphasizes informative samples via a dynamic-weighting loss. Together, these innovations enable more accurate, robust, and generalizable segmentation without relying on annotated data or external simulators.

\bibliographystyle{IEEEtran}
\bibliography{bib}

\newpage

\iftrue
\vspace{-0.5cm}
\begin{IEEEbiography}[{\includegraphics[width=1in,height=1.25in,clip,keepaspectratio]{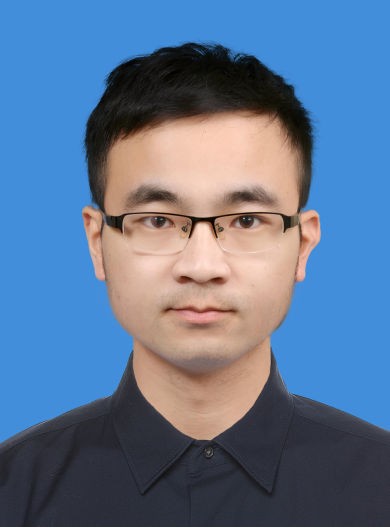}}]{Yifan Zhang}
	received the B.E. degree from the Huazhong University of Science and Technology (HUST), and the M.E. degree from Shanghai Jiao Tong University (SJTU), and the Ph.D. degree from the Department of Computer Science, City University of Hong Kong. Since 2025, he has been a lecturer at Shanghai University. His research interests include deep learning and 3D scene understanding.
\end{IEEEbiography}
\vspace{-0.5cm}

\begin{IEEEbiography}[{\includegraphics[width=1in,height=1.25in,clip,keepaspectratio]{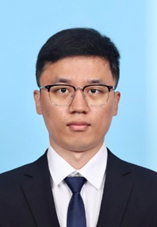}}]{Wei Zhang}
	received the Ph.D. degree in mechanical engineering from Shanghai Jiao Tong University, China, in 2023. He is now an assistant professor at School of Mechatronic Engineering and Automation, Shanghai University. His research interests cover perception, positioning, and navigation of intelligent agricultural robots and field robots.
\end{IEEEbiography}
\vspace{-0.5cm}

\begin{IEEEbiography}[{\includegraphics[width=1in,height=1.25in,clip,keepaspectratio]{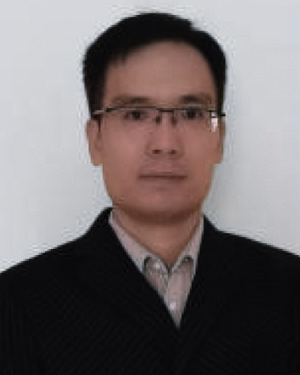}}]{Chuangxin He}
received the Ph.D. degree from Shanghai Jiao Tong University, Shanghai, China, in 2011. He has been an Associate Professor with the School of Mechatronic Engineering and Automation, Shanghai University, Shanghai. He has published more than 30 articles. His current research interests include underwater navigation, automation/intelligence/networking control of complex equipment, unmanned vehicles and robotics, and embedded systems with IoT technology.
\end{IEEEbiography}
\vspace{-0.5cm}

\begin{IEEEbiography}[{\includegraphics[width=1in,height=1.25in,clip,keepaspectratio]{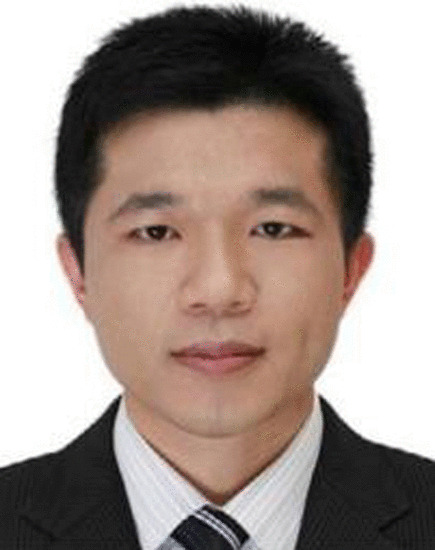}}]{Zhonghua Miao}
	received the Ph.D. degree in mechatronic engineering from Shanghai Jiao Tong University, Shanghai, China, in 2010.
	He is currently a Full Professor and Doctoral Advisor with the School of Mechatronic Engineering and Automation, Shanghai University, Shanghai, China. His research interests include intelligent robotics, measurement and control, fault diagnosis, and agricultural machinery systems.
\end{IEEEbiography}
\vspace{-0.5cm}

\begin{IEEEbiography}[{\includegraphics[width=1in,height=1.25in,clip,keepaspectratio]{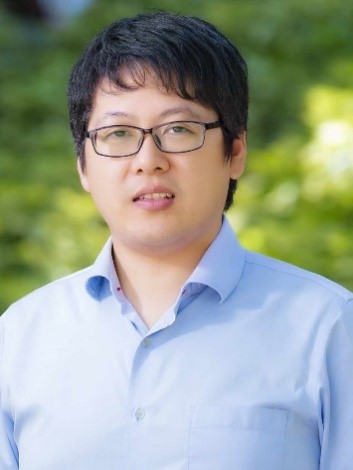}}]{Junhui Hou}
	(Senior Member, IEEE)  is an Associate Professor with the Department of Computer Science, City University of Hong Kong. He holds a B.Eng. degree in information engineering (Talented Students Program) from the South China University of Technology, Guangzhou, China (2009), an M.Eng. degree in signal and information processing from Northwestern Polytechnical University, Xi’an, China (2012), and a Ph.D. degree from the School of Electrical and Electronic Engineering, Nanyang Technological University, Singapore (2016). His research interests are multi-dimensional visual computing.
	
	Dr. Hou received the Early Career Award (3/381) from the Hong Kong Research Grants Council in 2018 and the NSFC Excellent Young Scientists Fund in 2024. He has served or is serving as an Associate Editor for \textit{IEEE Transactions on Visualization and Computer Graphics}, \textit{IEEE Transactions on Image Processing}, \textit{IEEE Transactions on Multimedia}, and \textit{IEEE Transactions on Circuits and Systems for Video Technology}.  
\end{IEEEbiography}
\vspace{-1.cm}

\fi

\vfill

\end{document}

%\title{Spatial-Temporal Graph Attention Network for Multi-Frame 3D Object Detection}
\title{Unsupervised Online 3D Instance Segmentation with
	Synthetic Sequences and Dynamic Loss \\ (Supplementary Materials)}

%缩写可以是: STEMD, STETR(3D), or STEP3D

%
%
% author names and IEEE memberships
% note positions of commas and nonbreaking spaces ( ~ ) LaTeX will not break
% a structure at a ~ so this keeps an author's name from being broken across
% two lines.
% use \thanks{} to gain access to the first footnote area
% a separate \thanks must be used for each paragraph as LaTeX2e's \thanks
% was not built to handle multiple paragraphs
%
%
%\IEEEcompsocitemizethanks is a special \thanks that produces the bulleted
% lists the Computer Society journals use for "first footnote" author
% affiliations. Use \IEEEcompsocthanksitem which works much like \item
% for each affiliation group. When not in compsoc mode,
% \IEEEcompsocitemizethanks becomes like \thanks and
% \IEEEcompsocthanksitem becomes a line break with idention. This
% facilitates dual compilation, although admittedly the differences in the
% desired content of \author between the different types of papers makes a
% one-size-fits-all approach a daunting prospect. For instance, compsoc 
% journal papers have the author affiliations above the "Manuscript
% received ..."  text while in non-compsoc journals this is reversed. Sigh.

\author{
	Yifan Zhang, Wei Zhang, Chuangxin He, Zhonghua Miao,
	Junhui Hou,~\IEEEmembership{Senior Member,~IEEE},
	% IEEE Publication Technology,~\IEEEmembership{Staff,~IEEE,}
	% <-this % stops a space
	%\thanks{Manuscript received April 19, 2021; revised August 16, 2021.}
	\thanks{Y. Zhang is with the School of Mechatronic Engineering and Automation, Shanghai University, Shanghai, China, and also with the Department of Computer Science, City University of Hong Kong, Hong Kong. E-mail: yfzhang@shu.edu.cn;}
	\thanks{W. Zhang, C. He, and Z. Miao are with the School of Mechatronic Engineering and Automation, Shanghai University, Shanghai, China. E-mail: {chuangxinhe@shu.edu.cn;wzhang23@shu.edu.cn;zhhmiao@shu.edu.cn;}}
	\thanks{J. Hou is with the Department of Computer Science, City University of Hong Kong, Hong Kong. E-mail: jh.hou@cityu.edu.hk.}
}

% note the % following the last \IEEEmembership and also \thanks - 
% these prevent an unwanted space from occurring between the last author name
% and the end of the author line. i.e., if you had this:
% 
% \author{....lastname \thanks{...} \thanks{...} }
%                     ^------------^------------^----Do not want these spaces!
%
% a space would be appended to the last name and could cause every name on that
% line to be shifted left slightly. This is one of those "LaTeX things". For
% instance, "\textbf{A} \textbf{B}" will typeset as "A B" not "AB". To get
% "AB" then you have to do: "\textbf{A}\textbf{B}"
% \thanks is no different in this regard, so shield the last } of each \thanks
% that ends a line with a % and do not let a space in before the next \thanks.
% Spaces after \IEEEmembership other than the last one are OK (and needed) as
% you are supposed to have spaces between the names. For what it is worth,
% this is a minor point as most people would not even notice if the said evil
% space somehow managed to creep in.

% The paper headers
\markboth{}%
{Shell \MakeLowercase{\textit{et al.}}: Bare Demo of IEEEtran.cls for Computer Society Journals}
\maketitle

% To allow for easy dual compilation without having to reenter the
% abstract/keywords data, the \IEEEtitleabstractindextext text will
% not be used in maketitle, but will appear (i.e., to be "transported")
% here as \IEEEdisplaynontitleabstractindextext when the compsoc 
% or transmag modes are not selected <OR> if conference mode is selected 
% - because all conference papers position the abstract like regular
% papers do.
\IEEEdisplaynontitleabstractindextext
% \IEEEdisplaynontitleabstractindextext has no effect when using
% compsoc or transmag under a non-conference mode.

% For peer review papers, you can put extra information on the cover
% page as needed:
% \ifCLASSOPTIONpeerreview
% \begin{center} \bfseries EDICS Category: 3-BBND \end{center}
% \fi
%
% For peerreview papers, this IEEEtran command inserts a page break and
% creates the second title. It will be ignored for other modes.
\IEEEpeerreviewmaketitle

%\section{Additional Experimental Results}
%\subsection{Analysis of Convergence and Sample Efficiency}
%\appendix

\section*{Datasets}
%\subsection{Datasets}
\textit{1) SemanticKITTI~\cite{behley2019semantickitti}:} The SemanticKITTI dataset is a large-scale LiDAR dataset for evaluating 3D semantic and instance segmentation models. Collected in urban environments in Germany using a 64-beam LiDAR at 10 Hz, it contains over 200,000 point cloud samples across 21 sequences. These sequences are aligned with RGB images, with each point cloud containing around 40,000 points. The dataset is split into 10 training sequences and one validation sequence (sequence 08). For the validation set, the evaluation is based on object instances belonging to several classes, including car, bicycle, motorcycle, truck, other vehicles, person, bicyclist, and motorcyclist.
%The validation sequence, in particular, is notable for its length, comprising 4071 consecutive LiDAR scans, making it ideal for testing algorithms in dynamic and extended real-world scenarios. This dataset serves as a valuable resource for the development of unsupervised and online instance segmentation models, given its rich variety of environments and object types.

%\textit{2) PandaSet:} The PandaSet dataset was collected using two different LiDAR sensors: the rotating Pandar64 and the solid-state PandarGT, both capturing data at 10 Hz. For our experiments, we focus on the high-density point clouds provided by the PandarGT sensor, which are more challenging and richer in detail. The dataset is split into training and validation sets based on GPS coordinates, with data from San Francisco used for training and the remaining data for validation. Unlike other datasets, PandaSet does not include ground-truth instance labels. To generate these labels, we combine semantic segmentation and object detection annotations, as outlined in the supplementary material. The subset used in this work, denoted as PandaSet-GT, contains data from the PandarGT, offering a high-density point cloud ideal for testing unsupervised and online 3D instance segmentation algorithms in complex urban environments.

\textit{2) PandaSet~\cite{xiao2021pandaset}:} The PandaSet dataset provides a large-scale collection of multi-modal sensor data captured in urban environments across San Francisco and Silicon Valley. It uniquely combines high-resolution LiDAR data from two different sensors: the Pandar64, a 64-channel rotating LiDAR, and the PandarGT, a solid-state LiDAR that offers image-like resolution with forward-facing scans.
%PandaSet includes annotations for 28 object detection classes (via 3D bounding boxes) and 37 semantic segmentation classes, enabling detailed scene understanding. The dataset contains 48,000 camera images and 16,000 LiDAR sweeps from over 100 dynamic driving sequences, each lasting 8 seconds. Additionally, it provides synchronized GPS/IMU data and full sensor calibration for accurate spatiotemporal alignment.
For our study, we focus on the PandarGT subset (PandaSet-GT), as it is similar to the Velodyne HDL-64E used in SemanticKITTI. We split the dataset using GPS metadata, with sequences from San Francisco reserved for training and sequences from other regions (such as Silicon Valley) used for validation.
%\todo{ Since PandaSet lacks instance-level annotations, we generate them by combining semantic segmentation labels with object detection bounding boxes, as described in the supplementary material.}
Since PandaSet lacks instance-level annotations, we generate the ground truth in the same way as in UNIT~\cite{sautier2024unit}.
%by combining semantic segmentation labels with object detection bounding boxes, as described in the supplementary material.

\textit{3) nuScenes~\cite{caesar2020nuscenes}:} The nuScenes dataset, captured in Boston and Singapore, uses a 32-beam LiDAR sensor to gather data at a frequency of 20 Hz. It consists of 850 driving sequences, with 700 used for training and 150 for validation. Each sequence lasts 20 seconds, and although training and inference utilize all 20 Hz LiDAR scans, evaluation metrics are computed using a subset of the validation scans, as annotations are provided at 2 Hz. The official panoptic annotations are used for evaluation, which include instances of 11 categories such as barriers, traffic cones, and trailers. 

\begin{figure}[t]
	\centering
	\includegraphics[width=0.4\textwidth]{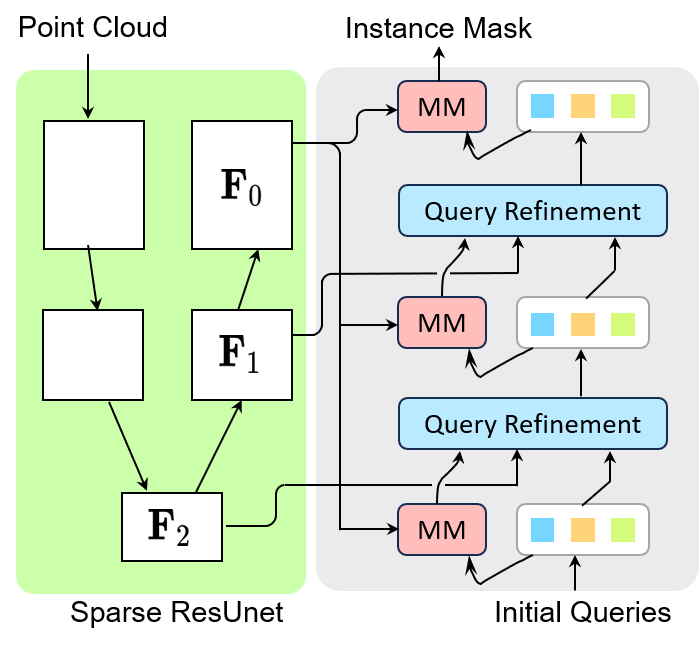} 
	\caption{
		Overview of class-agnostic Mask3D-based architecture~\cite{schult2022mask3d}.
%		 The backbone (green, left) is a sparse 3D U-Net that extracts multi-scale voxel features $(F_0, F_1, F_2)$ from the input point cloud (3D coordinates and intensity, voxelized at a fixed resolution).
		 The learned features are fed into a transformer decoder, where a set of learnable queries is iteratively refined through cross-attention with multi-scale features. At each refinement stage, a mask module (MM) computes similarity scores between queries and point features to produce instance masks. Since we operate in a class-agnostic setting, the semantic prediction head is omitted, and the network outputs only binary instance masks, ensuring consistency across the points belonging to the same object.
	}
	\label{fig:mask3d}
\end{figure}

\begin{figure*}[t]
	\centering
	\includegraphics[width=\textwidth]{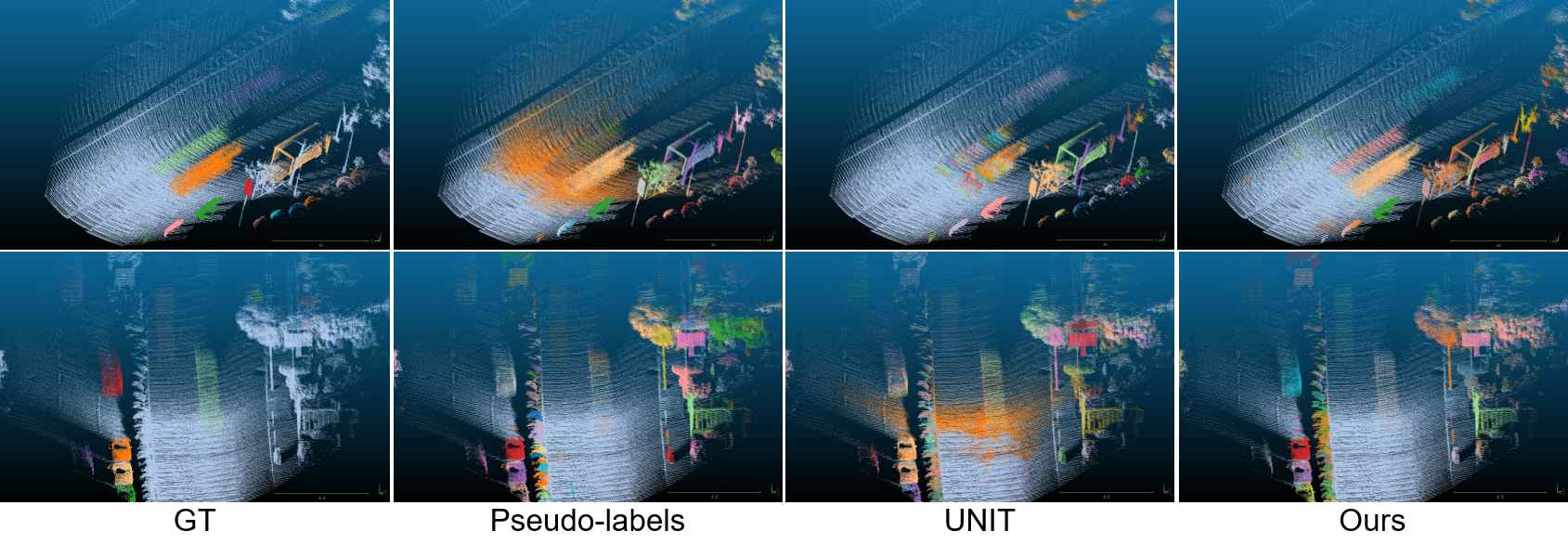} 
	\vspace{-0.1cm}
	\caption{
		%		Visual results on PandaSet-GT dataset. Best seen zoomed.
		\major{Visual results on PandaSet-GT dataset. Our method produces cleaner and more consistent instance masks compared with pseudo-labels and UNIT. Best viewed in zoom.}
	}
	\label{fig:vis_pandaset}
	\vspace{-0.1cm}
\end{figure*}

\begin{figure*}[t]
	\centering
	\includegraphics[width=\textwidth]{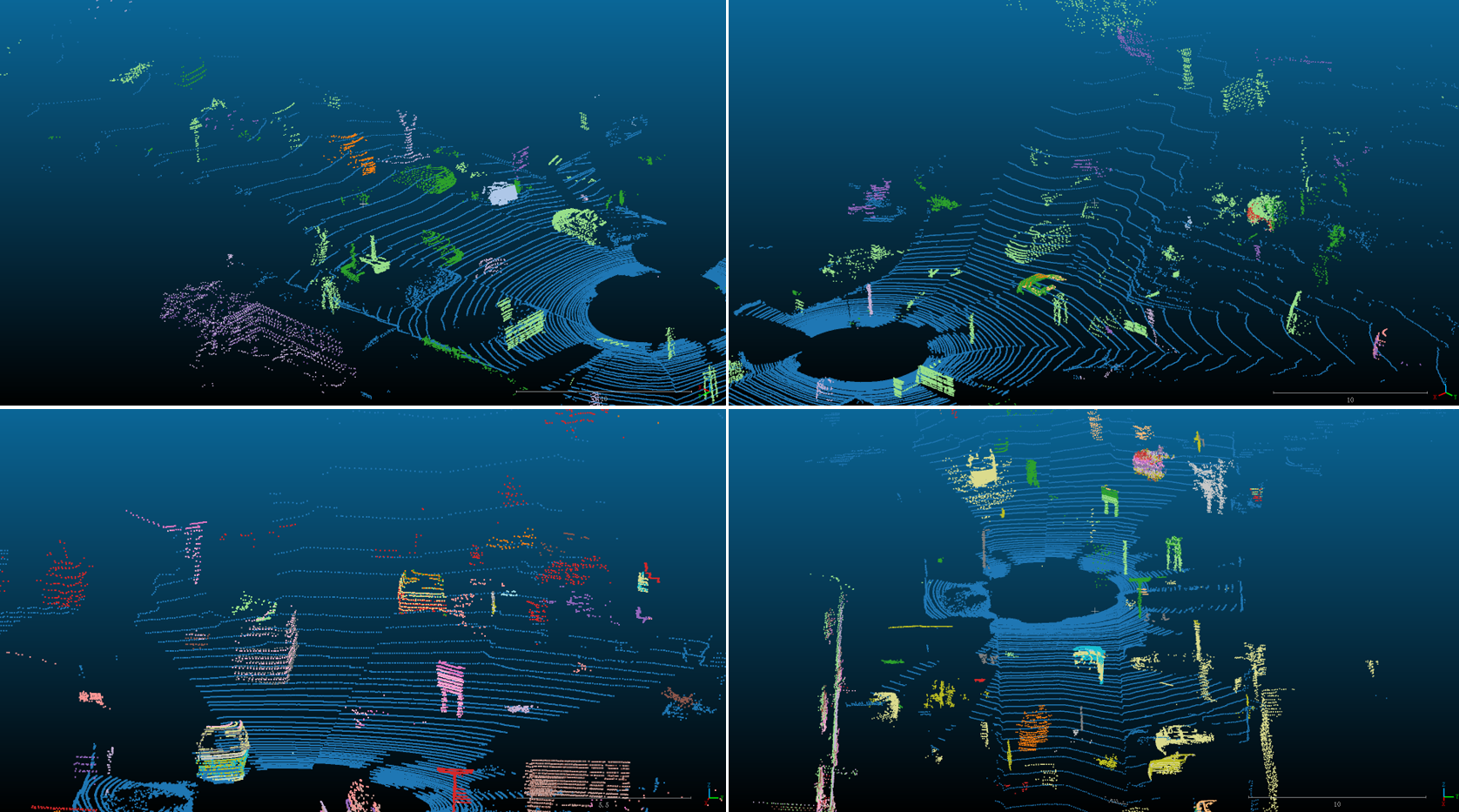} 
	\caption{
		Examples of synthesized point cloud sequences generated by our method. Each scene illustrates diverse object placements on realistic ground surfaces, ensuring spatial plausibility and variety for training.
	}
	\label{fig:syn_data}
\end{figure*}

\section*{Implementation Details}
%Our model architecture follows the design of [32], with a MinkUnet34 backbone [9] operating at a voxel resolution of 15 cm. To reduce computational overhead, we adopt mixed-precision training with bfloat16 and integrate flash attention [10] for efficiency. The network takes as input the LiDAR intensity values together with the Euclidean distance of each point from the sensor.
\textbf{Model Architecture.} Our model is built upon Mask3D~\cite{schult2022mask3d}, a transformer-based framework for 3D instance segmentation, but we adapt it to a class-agnostic setting by removing the semantic prediction head. As shown in Fig.~\ref{fig:mask3d}, the network directly predicts instance masks without assigning them to semantic categories. We employ a sparse 3D convolutional U-Net backbone based on the MinkowskiEngine to extract multi-scale point cloud features~\cite{choy20194d}. The input LiDAR scan is voxelized with a voxel size of 0.05 m (adjustable depending on the dataset), and each voxel encodes both the 3D coordinates and the intensity (reflectance) value of the contained points. The backbone outputs high-resolution voxel-wise features at multiple scales. For each query, we predict a binary mask over the input point cloud by computing the similarity between the query embedding and point features. Specifically, the dot product between the query and each point feature produces a response map, which is transformed into an instance mask by point-wise assignment. 
%At inference, the network outputs a set of masks covering all instances in the point cloud. Each input point is assigned to exactly one mask (query), yielding a dense class-agnostic instance segmentation.

\textbf{Training Protocol.} 
We employ a two-stage schedule following the existing online LiDAR segmentation framework~\cite{sautier2024unit}. The network is first optimized on individual scans using AdamW~\cite{loshchilov2017decoupled} with a cosine–annealed learning rate (initial value $1\times10^{-4}$), weight decay $1\times10^{-2}$, and batch size $3$. The number of epochs reflects the dataset scale: 50 for SemanticKITTI, 150 for PandaSet-GT, and 4 for nuScenes. During this stage, we apply light geometric perturbations—uniform scaling in $[0.9,1.1]$ and random rotations over $[0^\circ,360^\circ]$.
%First, we pretrain the network using single-scan inputs. This phase is optimized with AdamW~\cite{loshchilov2017decoupled}, employing a cosine-annealing learning rate schedule starting at $10^{-4}$, a weight decay of $10^{-2}$, and a batch size of 3. The number of epochs depends on the dataset size: 50 epochs for SemanticKITTI, 150 epochs for PandaSet-GT, and 4 epochs for nuScenes. In this phase, two forms of data augmentation are applied: random scaling within ±10\% and full 360° random rotations.
%These correspond to about 300k training iterations for SemanticKITTI and nuScenes, and roughly 200k iterations for PandaSet-GT.

%In the second phase, the network is fine-tuned on pairs of consecutive scans to incorporate temporal consistency. The optimizer and learning rate schedule remain unchanged, and we train for the same number of epochs as in the pretraining stage, with batch sizes of 4 for SemanticKITTI and 5 for both PandaSet-GT and nuScenes. Beyond the baseline setup, we integrate our key contributions during this stage: point cloud sequence synthesis is used to enrich the training data with diverse and realistic variations; flexible temporal sampling extends supervision to both adjacent and non-adjacent scan pairs, enabling the model to capture long-range temporal dependencies; and a dynamic-weighting loss emphasizes confident and informative dynamic instances while mitigating the influence of noisy or uninformative samples. In the temporal training stage, no augmentations are used except the synthetic point cloud data.

We then fine-tune on short sequences by feeding pairs of consecutive scans to enforce temporal coherence. Optimizer, schedule, and epoch counts are kept identical to Stage 1; the batch size is $4$ on SemanticKITTI and $5$ on PandaSet-GT and nuScenes. This phase integrates our three core components: (i) \emph{point-cloud sequence synthesis} to broaden the training distribution with diverse, physically plausible sequences. We generate synthetic sequences by inserting up to 600 objects per scan ($N_s=600$) sampled from a database of temporal object snippets. No additional augmentations are used in this phase beyond the injected synthetic sequences. (ii) \emph{Flexible temporal sampling} that randomly draws both adjacent and non-adjacent pairs (and their reversed order) to expose the model to long- and short-range dynamics, with a maximum gap of four frames. (iii) A \emph{dynamic-weighting loss} that up-weights confident, motion-informative instances while reducing the influence of noisy or uninformative ones. In the loss design, we employ a confidence-based scaling factor $w_{ij}$ with $\alpha=0.6$ and $\epsilon=0.1$ to down-weight uncertain predictions, and further weight each instance according to its centroid displacement across frames, normalized with a smoothing factor $\beta=0.2$. 

%For data augmentation, we synthesize sequences by inserting up to 600 objects per scan, sampled from a database of temporal object snippets. Placement is guided by a ValidMap built on ground points in the BEV plane, with collision detection and safety margins ensuring spatial plausibility, while each object is randomly rotated and slightly scaled before insertion.

% To incorporate temporal context, we adopt flexible temporal sampling, where anchor frames are paired not only with their immediate successor but also with frames up to four steps apart, with equal probability of reversing the order; this balances adjacent and non-adjacent pairs to expose the model to both short- and long-range dependencies. 

\section*{Additional Visual Results}
Figure~\ref{fig:vis_pandaset} presents qualitative comparisons on the PandaSet-GT dataset. Our method yields more accurate and temporally consistent instance segmentation compared with pseudo-labels and UNIT. 
%In nuScenes, UNIT introduces inconsistent predictions (highlighted with red circles), whereas our approach produces temporally consistent masks with clearer object boundaries. 
On PandaSet, which contains complex urban scenes with diverse object types, our method better preserves object shapes and reduces over-segmentation, leading to results that are closer to the ground-truth annotations. These improvements highlight the robustness and generalization of our approach across different datasets.

Figure~\ref{fig:syn_data} presents examples of the synthetic point cloud data produced by our proposed sequence synthesis pipeline. As shown, objects are sampled from an object database and placed on valid ground regions while avoiding collisions. This results in scenes that preserve realistic spatial arrangements and context. The generated sequences capture diverse object distributions, orientations, and densities, enriching the training data beyond what is available in raw scans. Such diversity is critical for improving model robustness and generalization in dynamic real-world environments.

% if have a single appendix:
%\appendix[Proof of the Zonklar Equations]
% or
%\appendix  % for no appendix heading
% do not use \section anymore after \appendix, only \section*
% is possibly needed

% use appendices with more than one appendix
% then use \section to start each appendix
% you must declare a \section before using any
% \subsection or using \label (\appendices by itself
% starts a section numbered zero.)
%

% \appendices
% \section{Notations}
% The notations used in this paper are summarized in Table~\ref{table:notations}.

% \setcounter{table}{0} % 将表格计数器重置为0

% you can choose not to have a title for an appendix
% if you want by leaving the argument blank
%\section{}
%Appendix two text goes here.

% use section* for acknowledgment
\ifCLASSOPTIONcompsoc
  % The Computer Society usually uses the plural form
%  \section*{Acknowledgments}
\else
  % regular IEEE prefers the singular form
%  \section*{Acknowledgment}
\fi

% The authors would like to thank...

% Can use something like this to put references on a page
% by themselves when using endfloat and the captionsoff option.
\ifCLASSOPTIONcaptionsoff
  \newpage
\fi

% trigger a \newpage just before the given reference
% number - used to balance the columns on the last page
% adjust value as needed - may need to be readjusted if
% the document is modified later
%\IEEEtriggeratref{8}
% The "triggered" command can be changed if desired:
%\IEEEtriggercmd{\enlargethispage{-5in}}

% references section

% can use a bibliography generated by BibTeX as a .bbl file
% BibTeX documentation can be easily obtained at:
% http://mirror.ctan.org/biblio/bibtex/contrib/doc/
% The IEEEtran BibTeX style support page is at:
% http://www.michaelshell.org/tex/ieeetran/bibtex/
\bibliographystyle{IEEEtran}
% argument is your BibTeX string definitions and bibliography database(s)
\bibliography{bib}